# Unsupervised representation learning
# with Hebbian synaptic and structural plasticity in
# brain-like feedforward neural networks

Naresh Ravichandran [a,*], Anders Lansner [a,b], Pawel Herman [a,c,*]

a.  Division of Computational Science and Technology, School of Electrical Engineering
    and Computer Science, KTH Royal Institute of Technology and Swedish e-Science
    Research Centre, Stockholm, Sweden
    b.  Department of Mathematics, Stockholm University, Stockholm, Sweden
    c.  Digital Futures, KTH Royal Institute of Technology, Stockholm, Sweden

* Corresponding authors.

{nbrav,ala,paherman}@kth.se

***Abstract***: Neural networks that can capture key principles underlying brain
computation offer exciting new opportunities for developing artificial intelligence and
brain-like computing algorithms. Such networks remain biologically plausible while
leveraging localized forms of synaptic learning rules and modular network
architecture found in the neocortex. Compared to backprop-driven deep learning
approches, they provide more suitable models for deployment of neuromorphic
hardware and have greater potential for scalability on large-scale computing clusters.
The development of such brain-like neural networks depends on having a learning
procedure that can build effective internal representations from data. In this work, we
introduce and evaluate a brain-like neural network model capable of unsupervised
representation learning. It builds on the Bayesian Confidence Propagation Neural
Network (BCPNN), which has earlier been implemented as abstract as well as
biophysically detailed recurrent attractor neural networks explaining various cortical
associative memory phenomena. Here we developed a feedforward BCPNN model to
perform representation learning by incorporating a range of brain-like attributes
derived from neocortical circuits such as cortical columns, divisive normalization,
Hebbian synaptic plasticity, structural plasticity, sparse activity, and sparse patchy
connectivity. The model was tested on a diverse set of popular machine learning
benchmarks: grayscale images (MNIST, F-MNIST), RGB natural images (SVHN,
CIFAR-10), QSAR (MUV, HIV), and malware detection (EMBER). The performance
of the model when using a linear classifier to predict the class labels fared
competitively with conventional multi-layer perceptrons and other state-of-the-art
brain-like neural networks.

plausible, representation learning, unsupervised learning, Hebbian plasticity, BCPNN
structural plasticity, cortical columns, modular neural networks, sparsity, rewiring,
self-organization.

# 1. Introduction

The human brain showcases an efficient and robust design for intelligent information processing that underlies sensorimotor processing, learning, memory, reasoning, planning, decision making, emotion, language, and much more. It excels over contemporary artificial intelligence in extracting and processing structure from real-time high-dimensional data and making cognitive inferences that produce complex behavior. This continues to be true despite the vast technological advances we have seen both in terms of computational power and speed, hardware, and algorithmic development as well as the availability of massive amounts of real-world data. Given the increasing abundance of parallelized high-performance computing resources along with the growing insights from cognitive, psychological and neuroscientific findings, there is now a surge of interest in the intersection of engineering approaches and brain sciences (Marblestone et al., 2016; Hassabis et al., 2017; Lake et al., 2017; Pehlevan and Chklovskii, 2019; Richards et al., 2019; Macpherson et al., 2021; Pulvermüller et al., 2021; Saxe et al., 2021; Zenke and Neftci, 2021). There are new opportunities to build intelligent systems based on a maturing "brain-like computing" paradigm that holds promise not only for artificial general intelligence but also related domains of machine learning, data science, robotics, computer vision and natural language processing among others. After all, brain-like approaches attempt to mimic a compact, robust, and power efficient system proven to work and excel at complex tasks of relevance for engineering domains listed above.

Artificial neural networks (ANNs) have been considered as a backbone of brain-like computing systems as they rest on a cognitive neurobiological premise that intelligent function emerges from the interactions of many simple nonlinear units with adaptive connection weights (Feldman and Ballard, 1982; McClelland and Rumelhart, 1986). Deep neural networks (DNNs; also called deep learning) as the most recent incarnation of ANNs have come to dominate the field largely due to their capability to discover efficient multi-layered representations of real-world data (Krizhevsky et al., 2012; Bengio et al., 2013; LeCun et al., 2015; Schmidhuber, 2015). Deep learning now achieves state-of-the-art performance in tasks involving visual object recognition, speech recognition, language translation, robotics, game playing, amongst several other application domains. The inspiration behind one of the most successful DNN architectures, namely convolutional neural networks (CNNs), partly originates from the functioning of the primate visual cortex, particularly the simple and complex cell receptive field properties of primary visual cortex (Hubel and Wiesel, 1968) together with the hierarchical stages of processing in the ventral visual stream for extracting increasingly complex visual representations (Fukushima and Miyake, 1982; Felleman and Van Essen, 1991; LeCun et al., 1998; Riesenhuber and Poggio, 1999). It is worth noting that DNNs, rather loosely derived from the workings of the brain's neuronal networks and largely supported with tremendous engineering efforts, have proved to consistently outperform previously established statistical machine learning methods and classical computer vision techniques having no relevance to neurobiology.

Although brain-like computing shares many objectives with deep learning, we emphasize that they are nevertheless different in architectural design, learning principles and thus also partly in functionality. The brain, in terms of cognitive function and neurobiological substrate provides, at best, broad inspiration but does not necessarily guide the development of such systems. Consequently, despite impressive performance in many complex large-scale data science problems they still struggle in some scenarios where human perception has no problem. For instance, DNNs are observed to be fragile to "adversarial examples" (like targeted white noise) that are imperceptible to humans (Szegedy et al., 2014; Goodfellow et al., 2015). DNNs also generalize poorly to natural distortions like occlusions and image contrast (Bowers et al., 2022; Geirhos et al., 2017; George et al., 2017; Malhotra et al., 2020). Furthermore, there are several problems in mapping deep learning computations to neurobiological mechanisms (discussed in Section 3). These problems question the brain-likeness of DNNs and their relevance to the brain-like computing paradigm despite their widespread success. We believe that many of the shortcomings of current DNNs can be overcome with brain-like neural network design that, on the one hand, accounts for relevant neurobiologically plausible mechanisms, and on the other hand, captures the core representation learning functionality of DNNs enabling them to serve as powerful pattern recognition systems.

Representation learning has received much attention in the DNN domain and reflects the general computational problem the brain's perceptual systems needs to solve. Real-world data, experienced through our sensory apparatus, rarely reflects the true structure and content of the world. The underlying causes generating the data offer a much simpler explanation of the data than the data itself (Bengio et al., 2013). For example, vision deals with data in the form of pixel intensities (or retinal cell voltages) but the underlying causes like object categories and their associated properties such as geometry, pose, texture, can be far more informative than the pixel intensities. The goal of representation learning should then be to discover the true statistical structure in the world by extracting these underlying causes of the data and to disentangle these representations (DiCarlo et al., 2012).

The approach we follow here builds on the Bayesian Confidence Propagating Neural Network (BCPNN) model of neocortical information processing (Lansner and Ekeberg, 1989; Lansner and Holst, 1996; Sandberg et al., 2002; Tully et al., 2014; Martinez et al., 2019; Ravichandran et al., 2020). The BCPNN adopts a discrete modular network architecture derived from the columnar organization of the primate neocortex: the network layer is modularized into hypercolumns, each in turn comprising several minicolumns locally competing in a soft-winner-take-all manner (Mountcastle, 1997; Douglas and Martin, 2004; Johansson and Lansner, 2007). Importantly, the neuronal and synaptic processes are framed as Bayesian probabilistic computations. The learning of connection weights is modelled as a Hebbian-Bayesian rule computing the co-activations of pre- and post-synaptic neurons in line with synaptic plasticity in the brain (Wahlgren and Lansner, 2001; Sandberg et al., 2002; Tully et al., 2014). Historically, the biophysically detailed spiking BCPNN models with a recurrently connected network architecture have been used to explain various forms of cortical associative memory (Fransén and Lansner, 1994; Fiebig and Lansner, 2017; Martinez et al., 2019; Fiebig et al., 2020; Chrysanthidis et al., 2022).

The primary objective of this work is to introduce the feedforward BCPNN model as a general-purpose algorithm for representation learning and to evaluate the model on a diverse range of datasets comparing to other brain-like computing and baseline deep learning models. Our novel contribution to the BCPNN model in this work includes (1) the structural plasticity algorithm for forming sparse patchy connectivity between layers in a data-driven manner, (2) creating an interface between continuous-valued data features to the BCPNN architecture using population vector coding methods and, (3) theoretical result showing the propagation and learning steps of the feedforward BCPNN model as equivalent to the E and M steps of the Expectation-Maximization algorithm for a multinomial (discrete) mixture model. We identified various brain-like attributes such as structural plasticity, sparse activity, sparse, patchy connectivity, and soft-winner-takes-all normalization which facilitate unsupervised learning of internal representations. At the same time, we emphasize that our abstract model leaves out many biophysical effects commonly incorporated in detailed computational neuroscience models (Lundqvist et al., 2011, 2013; Tully et al., 2014). We chose this abstract modelling framework to demonstrate functionality arising out of what we consider the core computational brain-like attributes of cortical information processing (O'Reilly, 1998; Pulvermüller et al., 2021) We further lay out our position on the attributes governing brain-like computing in Section 2.

## 2. What constitutes brain-like representation learning?

A staggering amount of brain data is being produced in anatomical, physiological, cognitive, behavioral, and clinical studies at different scales and levels of brain organization. Choosing a suitable level of abstraction and incorporating relevant details from such wealth of brain data for developing, implementing, and deploying brain-like systems poses a considerable challenge.  Here, we have attempted to identify brain-like attributes that are necessary and sufficient to enable robust representation learning capabilities while staying biologically constrained, i.e., avoiding any obvious violation of neurocomputational principles.

1. **Unsupervised learning**: The most successful form of deep learning today is based on supervised learning, where each sample in the dataset is accompanied by externally provided class label, and the network learns representations to correctly predict these labels (LeCun et al., 2015). The brain on the other hand, receives multimodal sensory data with hardly any explicit supervision. Representation learning in the brain is thus mostly unsupervised, where the network learns the structure of the world in terms of objects and their properties, rather than class labels (DiCarlo et al., 2012; Bengio et al., 2013).
2. **Local and Hebbian synaptic plasticity**: A biologically plausible way to change the strength of synaptic weights is to use only the information available locally at the synapse, i.e., the co-activations of the pre-synaptic and post-synaptic neuronal activity. Such associative mechanisms belong to the general class of learning algorithms called Hebbian synaptic plasticity, where the synaptic weights are strengthened by the repeated co-activation of pre- and post-synaptic neurons (Hebb, 1949).
3. **Online (incremental) updating**: The update rules governing neuronal and synaptic computations must be online, that is continuously updating the neural unit activities and synaptic weights as the input data is provided sample by sample (Sandberg et al., 2002; Gepperth and Hammer, 2016). This contrasts with the batch (or minibatch) mode, common in deep learning, where each weight update is derived from multiple data samples.
4. **Sparse distributed activities**: In the neocortex, the neural activities are highly sparse, i.e., at any point of time, only a tiny fraction (less than 1%) of neurons are firing (Lennie, 2003; Olshausen and Field, 2004; Quiroga et al., 2008; Wolfe et al., 2010; Barth and Poulet, 2012; Ahmad and Scheinkman, 2019; Yoshida and Ohki, 2020). Furthermore, a distinction has been made between local and distributed representations as a neural coding scheme implemented the brain (Hinton et al., 1986; Kanerva, 1988; Smolensky, 1988; Bowers, 2009). In the extreme local representation, each concept can be mapped to exactly one unit and one unit is activated by exactly one concept, sometimes referred to as a grandmother cell (Barlow, 1972; Gross, 2002; Quiroga et al., 2005). In contrast, in a distributed representation there is many-to-many mapping: each concept activates many units, and each unit is in turn activated by many concepts. Although local representations have the advantage of being simple and interpretable, they generally suffer from poor generalization and scaling. Distributed representation displays several interesting properties such as automatic generalization to unseen data, resilience to failure of individual units, high representational capacity, content-addressable memory, etc. The representation scheme in the brain more closely resembles that of a sparse distributed representation (McClelland and Rumelhart, 1985; Quiroga et al., 2008).

5. **Sparse and patchy connectivity:** The connectivity between neurons in the brain is extremely sparse. For the human neocortex containing neurons in the order of $10^{10}$, the number of synapses is estimated to be in the order of $10^{14}$. This is six orders lower compared to fully connected network which would contain $10^{20}$ synapses, i.e., only one in a million potential synaptic connections exist. The connectivity is also highly structured with overrepresentation of connections between spatially local regions and cortical areas (Felleman and Van Essen, 1991; Markov et al., 2014). The connectivity formed is also found to be "patchy", i.e., axons (originating from excitatory pyramidal neurons) branch a few times and then terminate in local spatial clusters, or patches, with thousands of synapses made on a local group of neurons (Houzel et al., 1994; Kisvárday et al., 1997; Binzegger et al., 2007; Martin et al., 2017). Network models with such sparse and patchy connectivity are important considerations for brain-like computing systems (Johansson et al., 2006; Meli and Lansner, 2013; Ahmad and Scheinkman, 2019).

Clearly, there are many other important biological attributes we have ignored in this study. For example, spiking neural networks play a prominent role in many computational neuroscience models and give rise to rich dynamical behaviors. However, the spike timing of cortical pyramidal neurons is known to be highly irregular *in vivo*, making the spiking generation process closely approximate a stochastic Poisson distribution (Softky and Koch, 1993; Shadlen and Newsome, 1998). These observations leave room for rate-based neuron models where the firing rate (the number of spikes per time window) codes for information. In fact, we expect brain-like systems to be flexible to switch between different levels of biological detail so that they could robustly operate either with or without spiking activation, as previously demonstrated (Ravichandran et al., 2023c, 2024)

As mentioned, the proposed brain-like approach to unsupervised representation learning builds on the feedforward BCPNN architecture. We are aware of the importance of the abundant recurrent cortical connectivity in the brain giving rise to complex temporal dynamics. Together with long-range top-down connections from other cortical areas, recurrent connections may perform relevant perceptual functions. However, experiments measuring response latency and receptive field organization suggest that the early perceptual/sensory processing and pattern recognition in the brain is largely feedforward in nature while the recurrent connections may perform other perceptual function such as attentional gating, pattern completion, temporal processing, etc. (Fabre-Thorpe et al., 1998; Lamme and Roelfsema, 2000; Angelucci and Bressloff, 2006; DiCarlo et al., 2012; van Bergen and Kriegeskorte, 2020; Ravichandran et al., 2023a, 2024).

# 3. Related Work

## 3.1 Unsupervised Hebbian-like learning

The simplest form of unsupervised learning of representations involves learning synaptic weights from a layer of input neurons to a single (hidden) neuron. Learning using Hebbian plasticity in the form of straightforward (linear) correlations of pre- and post-synaptic activities is unstable and weights can increase/decrease unbounded (Sanger, 1989; Turrigiano, 2008a). This is because Hebbian synaptic plasticity and neural activities form a positive feedback loop leading to a runaway effect. Many attempts over the years have used additional constraints on the correlation-based plasticity rules (Bienenstock et al., 1982a; Oja, 1982; Linsker, 1988; Lansner and Ekeberg, 1989; Sanger, 1989; Földiák, 1990; Bell and Sejnowski, 1995; Olshausen and Field, 1996). Such modifications not only render connection weights stable and bounded but also show that Hebbian-like learning can pick up statistical properties of the data and perform useful feature extraction. Oja (1982) demonstrated that if correlative Hebbian plasticity is modified with a forgetting term, neural units with linear activation function can learn the first principal component of the data (Oja, 1982). Subsequent work showed that the independent components of the data can be learned using Oja's rule modified with non-linear weight updates (Hyvärinen and Oja, 1998).

To represent different features from the data using many (>1) hidden neurons, the sparse coding model was developed where only a small fraction of neurons is active at any instant (Barlow, 1961; Földiák, 1990). This typically involves competition between neurons in the hidden layer. When Hebbian-like learning rules are coupled with sparse coding in the hidden layer, the network optimizes the mutual information between the input and hidden layer (Bell and Sejnowski, 1995). In the case of visual input, the weights learnt by the network resemble Gabor-like edge filters, similar to the receptive fields found in the simple cells of the primary visual cortex (Bell & Sejnowski, 1997; Olshausen & Field, 1996; Rehn & Sommer, 2007; Rozell et al., 2008; Zylberberg et al., 2011). Derived from the divisive normalization model of cortical circuits, competition between neurons has been modelled via reciprocal connections between excitatory (pyramidal) neurons and inhibitory neurons (basket cells, for example) (Fransén and Lansner, 1995; Carandini and Heeger, 2011; Isaacson and Scanziani, 2011). This divisive normalization model typically creates a soft-winner-takes-all competition between neurons than has been approximated by softmax normalization and used for promoting selectivity of features (Sandberg et al., 2002; Nessler et al., 2009, 2013; King et al., 2013; Moraitis et al., 2022).

While many of the above brain-like models rely on selected brain-like attributes and explain biological properties like receptive field organization and neuronal dynamics, their relevance for large-scale real-world pattern recognition on machine learning benchmarks has been limited. Models with spike-timing-dependent plasticity (using timing of first spike) have been used for learning features from image datasets, showing their usefulness for classification when coupled with MLP (Ferré et al., 2018) or support vector machines (Kheradpisheh et al., 2018a). In more recent work, several studies have addressed the problem on how to learn distributed representations using brain-like principles that can be scaled to machine learning benchmarks (Krotov and Hopfield, 2019; Miconi, 2021; Ravichandran et al., 2021; Journé et al., 2022; Lagani et al., 2022; Sa-Couto and Wichert, 2022). These works have demonstrated that the hidden layer with local synaptic learning can learn useful representations when coupled with local inhibition within the hidden layer to promote selectivity between the hidden neurons. Such models working with mostly brain-like attributes performed competitively on medium complexity machine learning datasets, MNIST and CIFAR-10. More recently, Journé et al. showed that networks with Hebbian plasticity using rectified exponential activations can perform well in multilayer settings and scale to big machine learning benchmarks such as STL-10 and ImageNet (Journé et al., 2022).

Illing et al., (2019) examined how neural networks with shallow architecture built with brain-like principles compete on machine learning benchmarks. For this, they compared many shallow unsupervised learning models with local learning rules. They found that the best performing models were *l*-PCA (Principal Component Analysis), *l*-ICA (Independent Component Analysis), *l*-RG (Random Gabor; no learning in this case), where "*l*" stands for spatially local receptive fields. They concluded that shallow networks could perform competitively on various small to medium complexity machine learning datasets (MNIST and CIFAR-10).

In the field of machine learning, representation learning is recognized as the key component for building efficient learning systems (Bengio et al., 2013; LeCun et al., 2015). Early research on unsupervised representation learning explored different architectures and learning rules that have relevance to the brain, such as autoencoders (AEs) (Hinton and Zemel, 1994; Hinton and Salakhutdinov, 2006) and Restricted Boltzmann Machines (RBMs) (Ackley et al., 1985). Both models train a layer of hidden neurons by reconstructing the input layer activities and use the reconstruction error as the learning signal. The RBMs and AEs can also be used for multilayer learning by stacking more layers on top of each other (Bengio et al., 2007, 2013; Erhan et al., 2010; Vincent et al., 2010).

## 3.2    Supervised biologically plausible backpropagation

Another thread of investigation into brain-like representation learning approaches constitute biologically plausible versions of deep learning. Recent years have seen growing interest in this approach, i.e., to reconcile the requirements of deep learning with neurobiologically plausible mechanisms (Marblestone et al., 2016; Bartunov et al., 2018; Illing et al., 2019; Pehlevan and Chklovskii, 2019; Richards et al., 2019; Whittington and Bogacz, 2019; Lillicrap et al., 2020; Payeur et al., 2021; Saxe et al., 2021; Zenke and Neftci, 2021). The success of deep learning has been predominantly attributed to the efficient use of the backprop (backpropagation of errors) algorithm for learning the weights in the network (Bengio et al., 2013; LeCun et al., 2015). Learning with backprop solves a more general credit assignment problem: determining how much each connection weight in the network should be increased (or decreased) in order to minimize a global loss function.

Although backprop has proven to be an extremely efficient algorithm in learning representations useful for a given task, there are several issues that make it an unlikely candidate model for synaptic plasticity in the brain. The most apparent issue with backprop is that the synaptic learning rule is neither local nor Hebbian. The backprop algorithm demands error gradient signals to be communicated from distant output layers. A straightforward mapping of this algorithm to the brain requires a forward phase where activity travels forward through all the layers of the network and a backward phase where error gradient signal travels backward through the network using the same connectivity, i.e., synapses reversing their direction of transmission with the post-synaptic neuron sending error signal back through the axon to the pre-synaptic neuron, a mechanism highly implausible to occur in the brain.

Attempts towards biologically plausible deep learning approaches typically start by replacing problematic elements in backprop-based DNNs with biologically plausible mechanisms while retaining representation learning functionality and performance. These approaches constitute a variety of models, but most commonly aim for attaining local synaptic plasticity. This is done by replacing the backward phase of backprop with explicit top-down synaptic connections which plays the role of carrying the error gradient signals required for credit assignment (Lillicrap et al., 2020). The forward and backward connections can also be integrated at the single neuron level, when complex neuron models with non-linear dendritic computation are used (rather than a point neuron model as assumed here) (Guerguiev et al., 2017; Sacramento et al., 2018; Payeur et al., 2021). These approaches make use of the observations that cortical pyramidal neurons have morphologically distinct apical and basal dendritic segments and model them to perform distinct computational functions of receiving bottom-up feedforward and top-down feedback signals respectively.

In other approaches, the credit assignment can also be done by using a global reward signal encoding the label prediction error and modulating the plasticity of all the connection weights (Roelfsema and Holtmaat, 2018; Illing et al., 2020; Pozzi et al., 2020). These models have been tested on standard machine learning datasets like MNIST, CIFAR-10, and ImageNet. Most of them have difficulty scaling to large real-world datasets (Bartunov et al., 2018). Even though these models mitigate some problems with backprop and use local synaptic learning rules, the degree of biological realism remain questionable. Most of them still require supervised learning, convolutional layers (with weight sharing mechanism, which is unbiological), and separate forward and backward passes through the network (Crick, 1989; Lillicrap et al., 2020).

## 3.3    Models chosen for comparison

We summarized the models discussed so far along with the feedforward BCPNN in Table 1 based on how they score on brain-like attributes. The first set of models compared are shallow networks with unsupervised learning and local learning rules: RBM, AE, *l*-ICA, *l*-PCA, *l*-RG (Illing et al., 2019). The second set of models are backprop models with supervised learning modified to have local synaptic learning: *l*-FA, *l*-DTP and SDTP (Bartunov et al., 2018; Lillicrap et al., 2020). The third set comprises methods that employ neither unsupervised learning nor biological attributes, such as supervised backprop-based MLP. Since we have focused on non-spiking (rate-based) neural networks in this study we did not include work on spiking neural networks performing representation learning (Diehl and Cook, 2015; Kheradpisheh et al., 2018b; Pfeiffer and Pfeil, 2018; Tavanaei et al., 2019; Taherkhani et al., 2020; Zenke and Neftci, 2021; Cramer et al., 2022; Ravichandran et al., 2023c, 2024).

*Table 1: Brain-like attributes of selected neural network models*

| Model type | Model name | Unsupervised learning | Local plasticity | Hebbian plasticity | Online learning | Sparse distributed activities | Sparse connectivity |
|---|---|---|---|---|---|---|---|
| Unsupervised Hebbian-like learning | Feedforward BCPNN | ✓ | ✓ | ✓ | ✓ | ✓ | ✓ |
| | RBM | ✓ | ✓ | ✓ | ✗ | ✗ | ✗ |
| | AE | ✓ | ✓ | ✗ | ✗ | ✗ | ✗ |
| | *l*-PCA | ✓ | ✓ | ✗ | ✓ | ✗ | ✓ |
| | *l*-ICA | ✓ | ✓ | ✗ | ✓ | ✗ | ✓ |
| | *l*-RG | ✗ | ✗ | ✗ | ✗ | ✗ | ✓ |
| | KH | ✓ | ✓ | ✗ | ✓ | ✓ | ✗ |
| Supervised Bio-backprop | BDSP | ✗ | ✓ | ✗ | ✓ | ✗ | ✓ |
| | *l*-DTP | ✗ | ✓ | ✗ | ✗ | ✗ | ✓ |
| | *l*-FA | ✗ | ✓ | ✗ | ✗ | ✗ | ✓ |
| DNN | MLP | ✗ | ✗ | ✗ | ✗ | ✗ | ✗ |

## 4. Feedforward BCPNN model

### 4.1 Network architecture

The feedforward BCPNN model contains two layers, input and hidden, connected in a feedforward fashion (Fig. 1A). The connections from the input to hidden layer perform core unsupervised representation learning by transforming the representations in the input layer to sparse distributed hidden (internal) representations. Distributed activities in both layers convey discrete-valued representations similar to previous non-spiking BCPNN models (Martinez et al., 2019; Ravichandran et al., 2020). The discrete-valued representation is inherent to the BCPNN model, and it derives from Bayes rule and from the discrete columnar organization observed in the primate neocortex (Mountcastle, 1957, 1997; Hubel and Wiesel, 1963, 1977; Douglas et al., 1989; Douglas and Martin, 2004).

In our model, we assume the cortical minicolumn as the fundamental computational unit (Fig. 1C). According to this abstraction, the cortical sheet constitutes several million cortical minicolumns, each comprising around 80 to 100 tightly interconnected neurons (Hubel and Wiesel, 1968; Mountcastle, 1997; Buxhoeveden and Casanova, 2002). The neurons within a minicolumn, stacked vertically within about a 40 μm diameter and perpendicular to the cortical surface, have been reported to exhibit functionally similar response properties (Mountcastle, 1997; Buxhoeveden and Casanova, 2002). Therefore, the minicolumns, rather than individual neurons, have been proposed as fundamental cortical computing units and have been considered to map to neuron-like units in abstract ANN models of pattern recognition and associative memory processing (Ballard, 1986; Alexandre et al., 1991; Lundqvist et al., 2006, 2011, 2013; DiCarlo et al., 2012).

The minicolumn units in our model are grouped into larger modules called hypercolumns (sometimes called macrocolumns; Fig. 1B). The activities of minicolumn units are normalized within each hypercolumn module through a local competitive mechanism rendering the hypercolumn to describe an attribute, i.e., the activity of each minicolumn unit represents the probability (confidence) of the feature value of the corresponding attribute. For example, a single hypercolumn could represent all possible orientations of edge stimuli within a small region of visual field, and each minicolumn would be then broadly tuned to a specific orientation, collectively covering the spectrum of 180 degrees (Hubel and Wiesel, 1963, 1977). This abstraction of the network architecture with hypercolumns of around 500 μm in diameter has been found to be repeated across many areas of the neocortex and has been proposed as a canonical cortical microcircuit (Mountcastle, 1997; Douglas and Martin, 2004). The local competition within each hypercolumn in the neocortex is typically mediated by reciprocal connections with shared inhibitory neurons (Carandini and Heeger, 2011; Isaacson and Scanziani, 2011). In summary, the BCPNN layer comprises $H$ hypercolumn modules, each consisting of M minicolumn units that collectively represent a data attribute mapped by the given hypercolumn. The activities of individual minicolumn units convey normalized discrete probabilities giving rise to discrete probability distributions within each hypercolumn for the corresponding data attribute. As a whole, they form a distributed representation of the input.

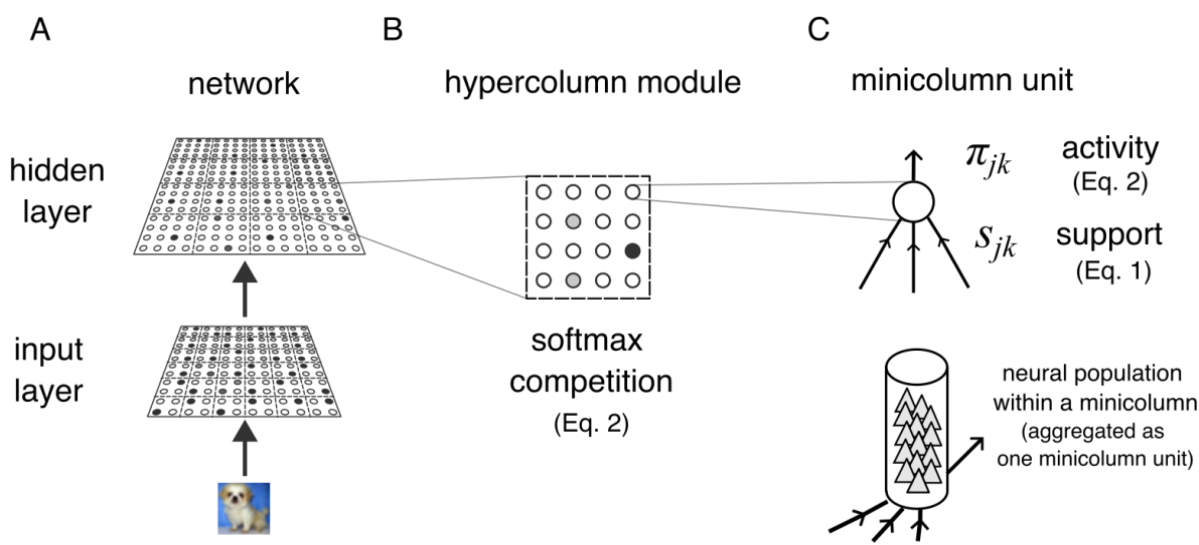

**Figure 1: Network schematic of the feedforward BCPNN model. A.** *The network contains two layers connected in a feedforward fashion. The raw data with continuous-valued attributes is transformed into a discrete-valued population vector representations in the input layer. The hidden layer performs the core unsupervised representation learning by transforming the representations in the input layer to hidden (internal) representations.* **B.** *Each layer is organized into hypercolumn modules constituting minicolumn units under softmax competition.* **C.** *The minicolumn unit acts as the fundamental computational unit in the model and corresponds (aggregates) to around 80-100 vertically stacked neurons (not simulated here) in the neocortex with functionally similar response properties.*

## 4.2 Unit activations

The activation function describes the activities of minicolumn in the hidden layer as a function of its synaptic input from the input layer. The input layer constitutes a set of $H_{inp}$ hypercolumns, each coding for an input attribute and denoted by the random variable $X_i$, $i \in \{1, \dots, H_{inp}\}$. The hypercolumn constitutes $M_{inp}$ minicolumns, each denoting a discrete feature value, $x_{im}$ ($m \in \{1, \dots, M_{inp}\}$) of the corresponding attribute, $X_i$. Similarly, the hidden layer constitutes $H_{hid}$ hypercolumns denoted by $Y_j$, $j \in \{1, \dots, H_{hid}\}$, each of which constitutes $M_{hid}$ minicolumns, $Y_j = y_{jk}$, $k \in \{1, \dots, M_{hid}\}$. The activity codes for the probability of discrete feature values in the data. We use the notation $\pi$ to refer to this probability. For the input layer, $\pi_{im} = P(X_i = x_{im})$, these activities are set by the data sample. When the input layer drives the hidden layer, the activation function computes the confidence probabilities, i.e., the posterior probability of a hidden minicolumn conditioned on the input layer activities, i.e., $\pi_{jk} = P(Y_j = y_{jk} \mid X_1, X_2, \dots, X_{H_{inp}})$.

The unit activation on the $n^{\text{th}}$ training iteration is given by:

$$s_{jk}^{(n)} = b_{jk}^{(n)} + \sum_{i=1}^{H_{inp}} \sum_{m=1}^{M_{inp}} \pi_{im}^{(n)} w_{imjk}^{(n)} c_{ij}^{(n)}, \tag{1}$$

$$\pi_{jk}^{(n)} = \frac{\exp s_{jk}^{(n)}}{\sum_{l=1}^{M_{hid}} \exp s_{jl}^{(n)}}. \tag{2}$$

Eq. 1 describes the computation of the support (summed synaptic input) received by the hidden minicolumn $y_{jk}$ by combining evidence from the input layer's activities. The term $b_{jk}$ refers to the bias of minicolumn unit $y_{jk}$, and the term $w_{imjk}$ refers to the connection weight between input minicolumn $x_{im}$ and hidden minicolumn $y_{jk}$. The activity $\pi_{jk}$ of the minicolumn $y_{jk}$ is computed from its support value by a softmax activation function (Eq. 2). The softmax function induces a soft-winner-take-all competition between the minicolumns within each hypercolumn module. This ensures the activities lie within the interval $[0, 1]$ and sum to one, thus each hypercolumn module follows a complete discrete probability distribution.

The term $c_{ij}$ is a binary connectivity variable indicating if there is an active ($c_{ij} = 1$) or silent connection ($c_{ij} = 0$) between the $i^{\text{th}}$ input hypercolumn and $j^{\text{th}}$ hidden hypercolumn. The connectivity matrix constituting $c_{ij}$ variables act as a binary mask that implement sparse patchy connectivity at the hypercolumn level, i.e., synapses between the minicolumns connecting a pair of hypercolumns are only active when the connectivity variable for the corresponding pair of hypercolumns is in active state. The connectivity matrix is updated adaptively by the structural plasticity mechanism (Section 4.4).

Interpreting unit activity as the probability (confidence) of the underlying feature is a key property of the BCPNN model. In the brain, such activities would correspond to the population-averaged firing rate of excitatory neurons within a cortical minicolumn (Fransén and Lansner, 1998; Lundqvist et al., 2006, 2011, 2013; Meyniel et al., 2015).

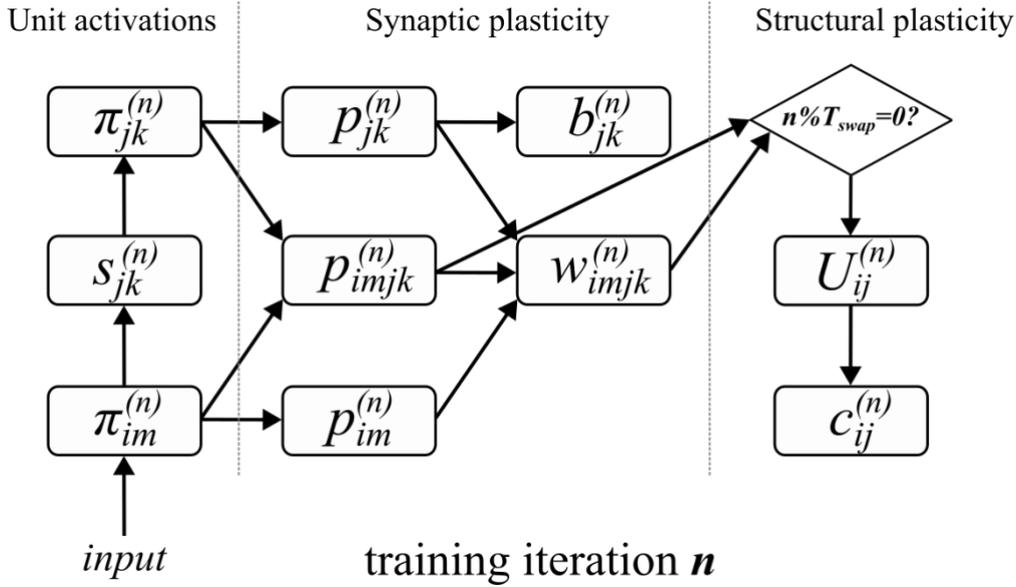

**Figure 2: Flowchart of the computations involved in a single training iteration of the feedforward BCPNN model.** *The input activity $\pi_{im}^{(n)}$ to each layer at training iteration $n$ is converted to support values $s_{jk}^{(n)}$, which in turn is converted to activity $\pi_{jk}^{(n)}$ through softmax normalization. The synaptic plasticity mechanism uses the pre- and post- synaptic minicolumn activities as input to compute p-traces. The structural plasticity mechanism operates every $T_{swap}$ iterations computing the usage term, $U_{ij}$, which in turn is used to compute the binary connectivity value $c_{ij}$.*

## 4.3   Synaptic plasticity

The synaptic plasticity rule determines the weights between pre-synaptic (input) and post-synaptic (hidden) minicolumns. The synaptic plasticity of the BCPNN framework is described concisely as Hebbian-Bayesian learning, since it converts probabilistic (Bayesian) estimation of occurrences and co-occurrences of events into a Hebbian form of synaptic plasticity (Lansner and Ekeberg, 1989; Wahlgren and Lansner, 2001; Sandberg et al., 2002; Tully et al., 2014). Each update in the Hebbian-Bayesian rule involves two steps: (1) computing memory traces local to each synapse that keeps track of the statistics of occurrences and co-occurrences of minicolumn activities and, (2) using the memory traces to determine the bias and weight parameters for each synaptic connection.

For each synapse indexed $imjk$, computation of memory traces involves incrementally (online) updating three $p$-traces: probability of pre-synaptic activity, $p_{im}$ (Eq. 3), probability of post-synaptic activity, $p_{jk}$ (Eq. 4), and joint probability of pre-synaptic and post-synaptic activities, $p_{imjk}$ (Eq. 5). On the presentation of the $n^{\text{th}}$ training sample, the $p$-traces between input minicolumn $x_{im}$ and hidden minicolumn $y_{jk}$ are updated as follows:

$$p_{im}^{(n)} = (1-\alpha)\,p_{im}^{(n-1)} + \alpha\,\pi_{im}^{(n)}, \tag{3}$$

$$p_{jk}^{(n)} = (1-\alpha)\,p_{jk}^{(n-1)} + \alpha\,\pi_{jk}^{(n)}, \tag{4}$$

$$p_{imjk}^{(n)} = (1-\alpha)\,p_{imjk}^{(n-1)} + \alpha\,\pi_{im}^{(n)}\,\pi_{jk}^{(n)}, \tag{5}$$

where $\alpha$ is the learning rate. The biases and weights are computed from the $p$-traces as follows:

$$b_{jk}^{(n)} = \log\ p_{jk}^{(n)}, \tag{6}$$

$$w_{imjk}^{(n)} = \log\ \frac{p_{imjk}^{(n)}}{p_{im}^{(n)}\ p_{jk}^{(n)}}. \tag{7}$$

The bias term (Eq. 6) encodes the pointwise self-information (or surprisal) of the post-synaptic minicolumn. Likewise, the weights encode the point-wise mutual information between the pre-synaptic and post-synaptic minicolumns. This allows computing the bias and weight parameters from memory traces that are completely spatiotemporally local to each synapse. As a crucial departure from traditional backprop-based deep learning models, this learning rule is online, local and, Hebbian.

The synaptic connections between the network layers can be realized in the cortex through long-range corticocortical synapses. Notably, individual weights may be excitatory (positive) or inhibitory (negative) and the latter are assumed to be mediated via di-synaptic inhibition with local inhibitory interneurons (Chrysanthidis et al., 2019), e.g., double bouquet cells inverting excitatory input to inhibition (DeFelipe et al., 2006).

Fig. 2 shows the flowchart with all the computations involved on the $n^{\text{th}}$ training iteration. Appendix A provides the theoretical derivation for the model, explaining how the feedforward BCPNN model performs probabilistic inference using the expectation-maximization algorithm through neuronal and synaptic computations. Appendix B provides the pseudocode for the unsupervised representation learning algorithm.

## 4.4    Structural plasticity

The anatomical structure of the brain's neural networks is highly dynamic, i.e., there is a continuous growth of new synapses and elimination of old synapses. Structural plasticity in the brain is a rewiring mechanism that changes the connectivity structure of the network by removing existing synaptic connections and creating new ones (Bailey and Kandel, 1993; Lamprecht and LeDoux, 2004; Stettler et al., 2006; Butz et al., 2009; Holtmaat and Svoboda, 2009). This continuous synaptic rewiring is, at least in part, an experience- and activity-dependent process (Butz et al., 2009; Holtmaat and Svoboda, 2009). Typically, structural plasticity operates on a slower timescale than synaptic plasticity, creating connections on which synaptic plasticity can change the efficacies of the existing connections. To design the algorithm to mimic structural plasticity, we incorporated three key experimental observations about connectivity and structural plasticity in the brain: (1) The fan-in (number of incoming synaptic terminals) is roughly constant across most neocortical (pyramidal) neurons as well as for each neuron across their lifetime (Pakkenberg and Gundersen, 1997; Alonso-Nanclares et al., 2008). (2) The neocortical connectivity is found to be highly structured and patchy, i.e., axonal connections originating from excitatory (pyramidal) neurons typically branch a few times and then terminate in local spatial clusters, or patches, with thousands of synapses made on a local group of neurons with a spatial extent of the same order as a cortical hypercolumn (Houzel et al., 1994; Kisvárday et al., 1997; Binzegger et al., 2007; Meli and Lansner, 2013). (3) Many of the incoming synaptic terminals on neocortical neurons are "silent", i.e., synapses that are present anatomically but do not allow synaptic transmission of spike events (Isaac et al., 1995; Liao et al., 2001; Kullmann, 2003; Kerchner and Nicoll, 2008).

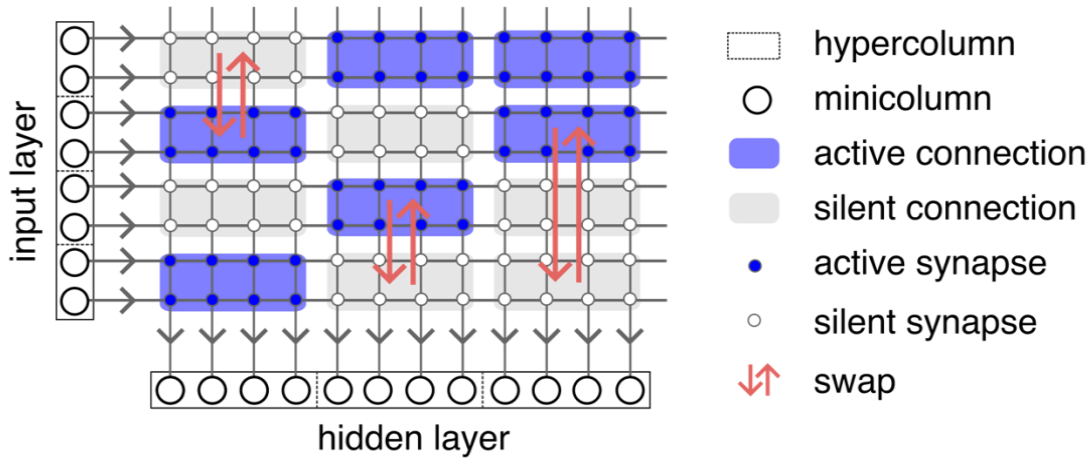

***Figure 3: Schematic of structural plasticity***. *BCPNN structural plasticity is a rewiring mechanism for adaptively rewiring the network connectivity. The connectivity matrix in the model between the input and hidden layer is always sparse and patchy, i.e., connections are always rewired for all minicolumns within each hypercolumn. Structural plasticity rewires the connections by silencing active connections with lower usage and activating silent connections with a higher usage. The existence of an active connection between the input and hidden layers is denoted by a blue patch, i.e., $c_{ij} = 1$ (and $c_{ij} = 0$ otherwise). The blue filled circles indicate active minicolumn-to-minicolumn synapses allowing propagation of activities.*

For modelling the structural plasticity mechanism, we define a binary variable $c_{ij}$ denoting the state of connection between the $i$th input and $j$th hidden hypercolumn (Fig. 3). As mentioned in Section 4.2, each hypercolumn-to-hypercolumn connection can be in one of two states, silent ($c_{ij} = 1$) or active ($c_{ij} = 0$), thereby masking synaptic connectivity between the constituent minicolumns in the spirit of patchy connectivity. The silent connections still take part in memory trace computations but not in synaptic transmission, while active connections participate in both. The role of the structural plasticity mechanism is to adaptively find a combination of active and silent connections that are effective at representation learning.

The inter-hypercolumn connections are initialized randomly from a binomial distribution with the constraint that all the hidden hypercolumns have the same number of active and silent incoming connections. The hyperparameter fan-in controls the overall level of sparsity in the connectivity. Over the training phase we compute for each connection between the $i$th input and $j$th hidden hypercolumn a "usage" term (Eq. 8), $U_{ij}$, that calculates the weight of each synapse connection (between constituent minicolumns) averaged by the joint probability $p_{imjk}^{(n)}$, indicating the average usage of the bundle of synaptic connections.

$$U_{ij}^{(n)} = \frac{\sum_{m,k} p_{imjk}^{(n)} w_{imjk}^{(n)}}{\sum_k c_{ik}^{(n)}}. \tag{8}$$

The numerator in the usage term is the mutual information between the input and hidden hypercolumns (substituting the weight term from Eq. 7) and the denominator is the fan-out of the input hypercolumn. The usage term thus computes a normalized pair-wise mutual information between the input and hidden hypercolumns for each connection. For each hidden hypercolumn, if the active connection with the lowest usage is smaller than the silent connection with the highest usage by a factor of at least $\rho$, the active connection becomes silent and simultaneously the silent connection becomes active (Eq. 9).

$$\forall j, \qquad c_{sj} \triangleq 1, \qquad c_{aj} \triangleq 0, \qquad \text{iff } U_{sj}^{(n)} > \rho\, U_{aj}^{(n)}, \tag{9}$$

$$\text{where } s = \operatorname*{argmax}_{i\,\in\,\text{silent}} U_{ij}^{(n)} \tag{10}$$

$$\text{and } a = \operatorname*{argmin}_{i\,\in\,\text{active}} U_{ij}^{(n)} \tag{11}$$

We call this operation a swap and use a hyperparameter $N_{swap}$ to set the number of swaps made per hidden hypercolumn per step of structural plasticity. Note that each swap retains the total number of incoming active and silent connections so the fan-in of all the hypercolumns is kept constant throughout. Each step of structural plasticity is executed with an interval of $T_{swap}$ training iterations. The structural plasticity mechanism in our model is correlation based, adaptive, sparse, and patchy, hence capturing many of the core principles from what has been known about neocortical connectivity (see Section 2).

## 4.5 Population vector coding in the input layer

The modularity of the BCPNN architecture with discrete minicolumns packed into hypercolumn modules imposes discrete-valued representations of data attributes. In the case when data is already discrete-value coded (e.g., black & white images), it can be fed directly to the input layer. However, several real-world datasets are continuous-valued (e.g., natural images, video, or speech data). Our input layer converts continuous-valued data to discrete-valued population vector code, and we tested two biologically plausible methods to achieve this: vector quantization with Gaussian Mixture Models (GMMs) and filtering with Difference of Gaussian (DoG) function.

### 4.5.1 Vector Quantization with Gaussian Mixture Models

Each attribute of data samples can be encoded by a population vector which quantizes the continuous-valued attribute using a discrete probability distribution over the corresponding feature values. For each continuous-valued attribute, we trained a GMM to obtain its discrete distribution (Neal and Hinton, 1998; Jain, 2010). All GMMs (for all the attributes) had $k$ mixture components (clusters), each with a parameter determining the mean of the Gaussian distribution and a covariance matrix set to the identity matrix. After fitting GMMs, for each attribute in an input data sample, the corresponding GMM outputs a vector of activities representing the posterior probabilities (memberships) of the data sample over the respective mixture components. This vector served as the activities of minicolumns in the hypercolumn encoding the given attribute. Although other clustering methods, e.g., $k$-means, could be used for that purpose, GMMs provide a probabilistic interpretation of the data in line with the BCPNN framework.

### 4.5.2 Difference of Gaussian filtering

Since the encoding of continuous-valued data into discrete-coded activities is a consequence of modular organization where groups of discrete minicolumns form hypercolumn modules, a similar computation can be expected to exist in the visual pathway from retina leading to the neocortex.

Retinal photoreceptors have graded voltage response directly reflecting the number of photons hitting the rod or cone cells. The early visual system needs an intermediate step to convert this continuous-valued retinal activities before projecting to the primary visual cortex with discrete-valued orientation columns. Lateral geniculate nucleus (LGN) located in the thalamus acts as this intermediate step, constituting cells with center-surround receptive field organization (Kuffler, 1953; Lettvin et al., 1959; DeAngelis et al., 1995). The LGN cells (as well as retinal ganglion cells) respond to a small approximately circular piece of the visual field with the center and surround regions of the receptive field acting antagonistically, i.e., one increasing while the other decreases the cell's response (DeAngelis et al., 1995). The LGN cells, as a whole, provide full coverage of the visual field (although with varying degree of density across the visual field). Furthermore, there is typically a complementary coverage. In other words, for a specific receptive field there are usually both on-center/off-surround and off-center/on-surround cells (DeAngelis et al., 1995). Many LGN cells display color opponency in addition to the grayscale on-center/off-surround receptive fields (Livingstone and Hubel, 1988; Dacey, 2000; Chatterjee and Callaway, 2003). Such cells show red/green opponency (red-center/green-surround and green-center/red-surround) as well as blue/yellow opponency (blue-center/yellow-surround and yellow-center/blue-surround).

Hence, our DoG filtering approach parallels more closely a biological visual system by incorporating well known properties of the receptive field organization. The standard way of modelling the center-surround receptive fields is by using DoG filters (Marr and Hildreth, 1980; DeAngelis et al., 1995; Lindeberg, 2013), where the images filtered with a small Gaussian filter (low standard deviation) is subtracted from the corresponding image filtered with a large Gaussian (high standard deviation). Computationally, DoG filtering performs edge detection and contrast enhancement on the images (Marr and Hildreth, 1980). We use an input layer with connections weights from the image data set as DoG filters with aperture sizes of $3 \times 3$ (small Gaussian) and $5 \times 5$ (large Gaussian). The responses are then normalized with a non-linear sigmoidal activation function. In the case of RGB images we used red/green and blue/yellow DoG filters in addition to the original DoG filters.

# 5. Experimental setup

The aim of our experiments was two-fold: (1) to investigate the representation learning capability of the feedforward BCPNN model by assessing the internal representations and, (2), to evaluate the classification performance of the model against other brain-like models over a diverse range of datasets.

## 5.1 Datasets

We chose several prominent machine learning benchmarks for our experimental evaluation. We worked with four image datasets each containing ten classes of objects: grayscale images of hand-written digits called MNIST (LeCun et al., 1998), grayscale images of fashion products called F-MNIST (Xiao et al., 2017), RGB images of street view house numbers called SVHN (Netzer et al., 2011), and RGB images of real-world objects called CIFAR-10 (Krizhevsky and Hinton, 2009). In addition to image data, we chose datasets where there is no known topology describing the relationship between data attributes. For this, we used a large malware detection dataset called EMBER containing attributes extracted from potentially malicious Windows portable executable files (Anderson and Roth, 2018) with the task of classifying the files as malicious or benign. We further used two Quantitative Structure-Activity Relationship (QSAR) datasets, the HIV and MUC datasets, where the chemical or biological activity of a molecule is to be predicted from its molecular description and properties (Goh et al., 2017; Wu et al., 2018). The HIV dataset represents a single-task classification problem and the MUC – a multi-task classification problem. For this, we encoded the modular description data as ECFP (Extended Connectivity Fingerprints) features and used it as input to our model. A summary of all the datasets used with the relevant details is listed in Table 2.

*Table 2: Summary of datasets used for evaluation.*

| Dataset | Data type | Task | # training samples ($N_{train}$) | # test samples ($N_{test}$) | # input attributes | # output classes |
|---|---|---|---|---|---|---|
| MNIST | Images (grayscale) | Classification | 60000 | 10000 | 28x28 | 10 |
| F-MNIST | Images (grayscale) | Classification | 60000 | 10000 | 28x28 | 10 |
| SVHN | Images (RGB) | Classification | 73257 | 26032 | 32x32x3 | 10 |
| CIFAR-10 | Images (RGB) | Classification | 50000 | 10000 | 32x32x3 | 10 |
| EMBER | Malware | Classification | 900000 | 200000 | 2381 | 2 |
| HIV | QSAR | Classification | 37722 | 4191 | 1024 | 2 |
| MUV | QSAR | Multi-task classification | 83413 | 9312 | 1024 | 12x2 |

## 5.2 Evaluation using a linear classifier

The model's hidden representations are expected to encode salient information from the data that would be useful for further downstream tasks. Although the feedforward BCPNN model does not require any label information for carrying out representation learning, we exploited the labels after the representations have been extracted to quantify the class separability and to compare with other machine learning models. For this, we used a simple linear classifier with softmax output units, trained on the hidden representations to predict the labels, as in recent studies on unsupervised deep learning models (Oord et al., 2018; Chen et al., 2020) and other brain-like neural networks (Bartunov et al., 2018; Krotov and Hopfield, 2019; Ravichandran et al., 2021; Moraitis et al., 2022).

For training the classifier, we used a cross-entropy loss function and minibatches of 100 samples with 25 training epochs (one epoch denotes one full sweep through the training samples). We used the Adam optimizer to set the learning rate schedule with parameters $\alpha = 0.001$, $\beta_1 = 0.9$, $\beta_2 = 0.999$, and $\epsilon = 1e - 7$ as originally defined (Kingma and Ba, 2015). This setup was identical for all models under comparison, thus ensuring the representation learning was evaluated uniformly. Since this classifier layer is not a part of the BCPNN network architecture, we did not aspire to preserve biological plausibility here but merely use it as a tool to evaluate the class separability of the representations generated.

## 5.3 Related models under comparison

For comparison, we grouped models into four categories based on their brain-like attributes (see Table 1), or lack thereof. For the first category, we used a linear classifier (as a baseline model) directly trained on the input data to predict class labels. For the second category, we sought out models with unsupervised Hebbian-like learning. For this, we used RBM and AE, as well as *l*-PCA, *l*-ICA, and *l*-RG, which were found to perform well on image recognition tasks based on previous work (Illing et al., 2019). For both RBM and AE, we used a hidden layer constituting 3000 binary units each. For training the AE, we used the binary cross-entropy loss function with minibatches of 20 samples and an additional L1 loss on the hidden unit activities to encourage sparse overcomplete representations. For training RBM, we used contrastive divergence (CD-*k*) with $k = 2$ steps of Gibbs sampling. We included the model developed by Krotov and Hopfield, which was found to perform well on MNIST and CIFAR-10 dataset (Krotov and Hopfield, 2019). For the third category, we chose backprop-based DL models using supervised learning that were modified with the goal of having local learning rules. For this we included BDSP, *l*-DTP, and *l*-FA (Bartunov et al., 2018; Lillicrap et al., 2020). For the fourth category, we included MLP, a simple backprop-based neural network with one hidden layer trained by supervised learning.

## 5.4 Implementation details

We ran the network simulations on a single GPU (NVIDIA A100) with 40 GB HBM2 memory. The feedforward BCPNN model was implemented in C++ with OpenACC for GPU parallelization. We used online updates (without batching data samples) in both the training and evaluation phases. The full list of parameters used is given in Table 3. The total (training + test) run time for small networks with hidden layer sized $H_{hid} = 30$, $M_{hid} = 100$, on MNIST dataset was approximately 3 minutes, while for larger network with $H_{hid} = 400$, $M_{hid} = 100$, each run took approximately 30 minutes. For SVHN and CIFAR-10 dataset, we pre-processed the data (using one of the two population vector coding methods described in Section 4.5) and created the input layer population-vector activities in Python 3.11 using the scikit-learn library. For QSAR datasets MUV and HIV, we used the MoleculeNet Python library for computing the ECFP (Extended Connectivity Fingerprints) features from the molecular description data (Wu et al., 2018). Among the related models under comparison, we implemented RBM, AE, and MLP in Python with TensorFlow v2.9 and ran on the same single GPU hardware.

*Table 3: Model parameters for the feedforward BCPNN network on MNIST*

| Type | Parameter | Value | Description |
|------|-----------|-------|-------------|
| Network architecture | $H_{inp}$ | 784 | # hypercolumns in input layer |
| | $M_{inp}$ | 2 | # minicolumns per hypercolumn in input layer |
| | $H_{hid}$ | 30 | # hypercolumns in hidden layer |
| | $M_{hid}$ | 100 | # minicolumns per hypercolumn in hidden layer |
| | $fanin$ | 78 | # active incoming connections per hypercolumn |
| Learning parameters | $\alpha$ | 0.0001 | Learning rate of *p*-traces |
| | $\lambda$ | 0.001 | Amplitude of Gaussian noise added to support term |

| | $N_{usup}$ | 5 | # epochs for unsupervised learning |
|---|---|---|---|
| Structural plasticity | $N_{swap}$ | 100 | # swaps to be performed per structural plasticity step |
| | $T_{swap}$ | 500 | # training iterations per structural plasticity step |
| | $\rho$ | 1.1 | Threshold between active and silent usage for swap |

# 6. Results

## 6.1    Characterizing the internal representations of BCPNN model

We assessed the internal representations generated by the hidden layer of the feedforward BCPNN model and characterized them quantitatively. For this, we trained the model ($H_{hid} = 30$, $M_{hid} = 100$, $fanin = 78$) on the MNIST dataset for $N_{usup} = 5$ epochs. We computed the histogram of the hidden layer activities (Fig. 4A) aggregated over all minicolumns and hypercolumns in the hidden layer and over test data samples. There was a strong tendency for activations to follow a sparse distribution, as most of the unit activities were around 0 and a smaller proportion around 1 (notice the logged y-scale), while intermediate values were mostly absent. Since the activities within each hypercolumn followed a discrete probability distribution, this suggested that collectively hypercolumns display high "confidence", i.e., winner-take-all dynamics. We quantified this winner-take-all property by computing the entropy of each hypercolumn averaged across the layer and across the test samples ($N_{test}$, indexed by $n$) as follows:

$$\bar{S}_{activity} = -\frac{1}{N_{test}H_{hid}} \sum_{n=1}^{N_{test}} \sum_{j=1}^{H_{hid}} \sum_{k=1}^{M_{hid}} \pi_{jk}^{(n)} \log \pi_{jk}^{(n)} \qquad (12)$$

For all the datasets (Table 4) $\bar{S}_{activity}$ was considerably low, below 0.7. For comparison, a hidden layer with all hypercolumns following a winner-takes-all distribution (minimum entropy) produced $\bar{S} = 0$ and a hidden layer with all hypercolumns following a uniform distribution (maximum entropy) produced $\bar{S} = 4.60$. The model's hidden representations thus closely approximated a winner-take-all distribution. Note that this behavior was not enforced by any BCPNN update equations; the softmax activation function of the minicolumns (Eq. 2) produces continuous-valued activities between 0 and 1. The network, through the learning mechanism, generates activities that closely approximate a winner-take-all distribution. The representations comply with the requirements of the BCPNN framework, displaying discrete coding with sparse activities.

Next, we measured the average "usage" of the minicolumns, which is different from the distribution of input-driven feedforward activations (posterior probabilities, $\pi$). Here we calculated the histogram of $p_{jk}$, i.e., marginal probabilities of the minicolumns aggregated across the layer, which indicates if the minicolumn has been activated across the data samples (Fig. 4B). The distribution peaked at approximately 0.01, which is the entropy of uniform distribution (maximum entropy), $p_{jk} = 1/M_{hid}$ with $M_{hid} = 100$. We quantified this by computing the entropy of the $p_{jk}$ values for each hypercolumn and averaged across the hidden layer, as follows:

$$\bar{S}_{usage} = -\frac{1}{H_{hid}} \sum_{j=1}^{H_{hid}} \sum_{k=1}^{M_{hid}} p_{jk}^{(n)} \log p_{jk}^{(n)} \tag{13}$$

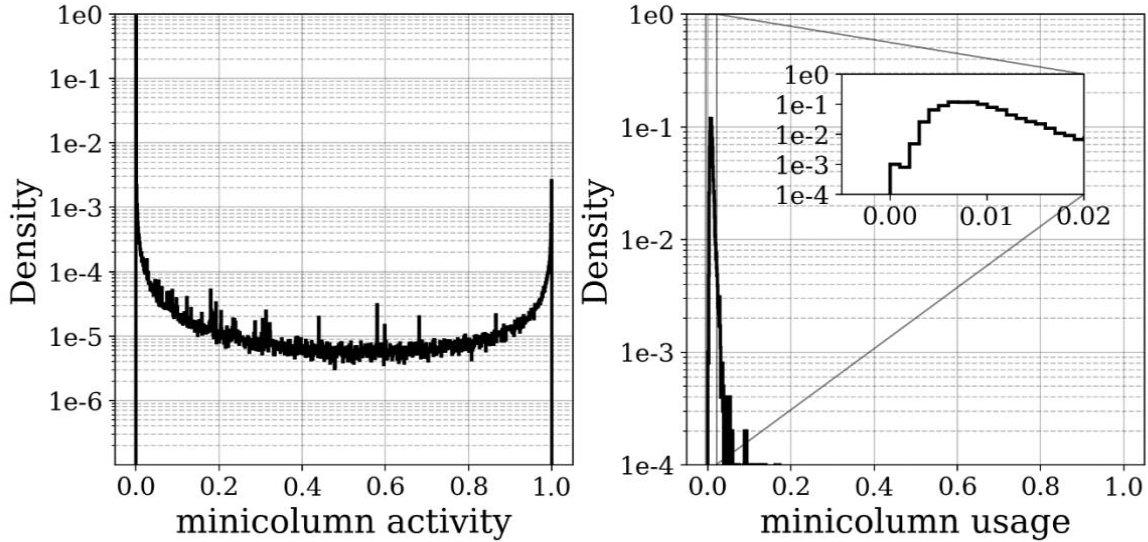

*Figure 4: Assessing the internal representations of feedforward BCPNN model trained on MNIST data. **Left.** Histogram of minicolumn activations (density) in the hidden layer (bin size of 0.001) shows that the activations peaked at 0 and at 1, and are scarce at intermediate values, closely approximating a winner-take-all distribution. **Right.** Histogram of minicolumn (marginalized) probabilities in the hidden layer indicating the usage of the units (bin size of 0.001). The probabilities are peaked at approximately 0.01 (inset figure), indicating an approximately uniform distribution where all minicolumns have uniform usage across the dataset.*

Across all the datasets we tested (Table 4), $\bar{S}_{usage}$ for the hidden layer activities closely approximated a hidden layer with all hypercolumns following a uniform distribution (maximum entropy), $\bar{S} = 4.60$. This showed that the usage of the minicolumns was tightly regulated with all the minicolumns equally activated across the samples while avoiding dead units, despite the activities being approximately winner-take-all per sample.

*Table 4: Entropy of minicolumn activities, $\bar{S}_{activity}$, and usage, $\bar{S}_{usage}$.*

| Dataset | $\bar{S}_{activity}$ | $\bar{S}_{usage}$ |
|---------|---------------------|-------------------|
| MNIST | $0.250 \pm 0.002$ | $4.416 \pm 0.005$ |
| F-MNIST | $0.646 \pm 0.012$ | $3.770 \pm 0.014$ |
| SVHN | $0.354 \pm 0.037$ | $3.418 \pm 0.013$ |
| CIFAR-10 | $0.570 \pm 0.002$ | $3.842 \pm 0.003$ |

A key property of representation learning is that the hidden layer activities encode the underlying factors that generated the data. Ideally, this requires the representations to be orthogonalized, i.e., activities produced from similar underlying factors are similar to each other and dissimilar from different underlying factors. To show how the representations are transformed from the input to hidden layer, we computed the pair-wise similarity matrix between the representations produced by the feedforward BCPNN model trained on MNIST and F-MNIST data. We used 1000 randomly chosen test data samples and computed cosine similarity ($S_c$) between all pairs of input layer activities (raw data) and between all pairs of the hidden layer activities. Fig. 5A and Fig. 5B show the similarity matrices, sorted by class labels, for the input and hidden layer activities, respectively, when trained on MNIST data. Fig. 5C and Fig. 5D show the corresponding plots when trained on the F-MNIST data. Considering that the class label is a key underlying factor of the data, the cosine similarity for samples from the same class should be close to 1 (high overlap between representations), and close to 0 for samples from different classes (minimal overlap between representations). The similarity matrix for the input data showed class ambiguity as samples were similar to other samples within and outside the class label, and the cosine scores were roughly around 0.5 for most pairs indicating considerable overlap. The similarity matrix for the feedforward-driven hidden representations were markedly different with most of the cosine scores close to zero (because the activations are sparse) and just a few non-zero scores occurred within the same label (small square patches along the diagonal in Fig. 5B and 5D).

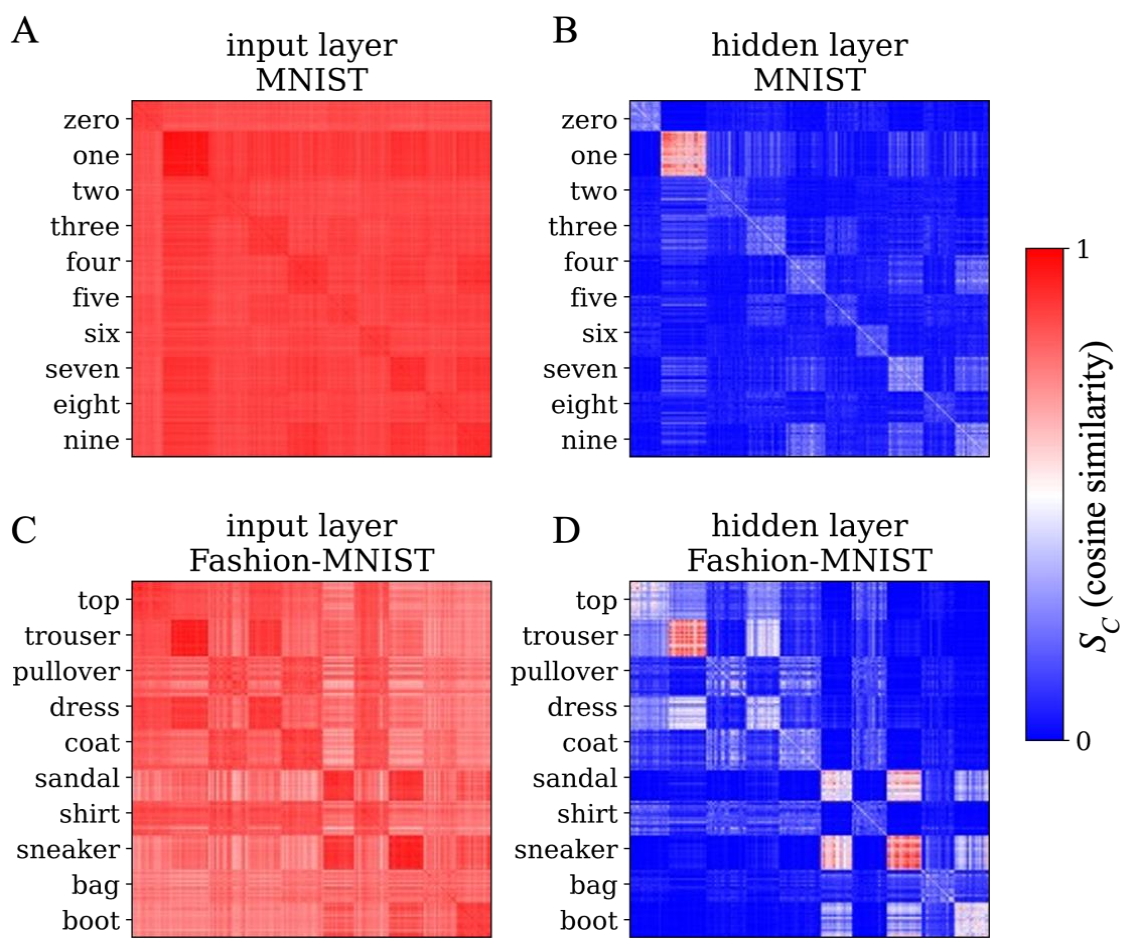

*Figure 5: Orthogonalization of representations by feedforward BCPNN model trained on MNIST and F-MNIST data. Pair-wise cosine similarity matrix of different representations sorted by labels for A. MNIST data. B. hidden layer representations driven by MNIST data, C. F-MNIST data, D. hidden layer representations driven by F-MNIST dataset.*

We quantified this by calculating a class-dependent similarity ratio, $B$, as the ratio of average similarity of activities between samples within the same class and average similarity between samples across all classes, as follows:

$$B = \frac{\sum_{u,v \in same\ class} S_c\left(\pi^{(u)}, \pi^{(v)}\right)}{\sum_{u,v \in all\ pairs} S_c\left(\pi^{(u)}, \pi^{(v)}\right)} \ , \tag{14}$$

For the model to have produced class-dependent similarity requires $B > 1$, and higher magnitude implies higher degree of class-dependent similarity. Across all the datasets we tested (Table 5), $B$ for the hidden layer activities was higher than the input data. The pair-wise similarities matrix as well as the class-dependent similarity ratio suggested that the feedforward BCPNN model did indeed orthogonalize the data in terms of class labels.

*Table 5: Class-dependent similarity ratio, B, for input data and hidden representations.*

| Dataset | $B_{inp}$ | $B_{hid}$ |
|---------|-----------|-----------|
| MNIST | 1.37 | $3.33 \pm 0.06$ |
| F-MNIST | 1.28 | $2.90 \pm 0.03$ |
| SVHN | 1.04 | $1.25 \pm 0.01$ |
| CIFAR-10 | 1.05 | $1.28 \pm 0.08$ |

## 6.2 Formation of receptive fields through structural plasticity

In neuroscience the receptive field of a neuron is defined as the region of sensory stimulus space that selectively modulates the activity of the neuron (Kuffler, 1953; Hubel and Wiesel, 1962, 1968). For instance, in primary visual cortex, neurons within each hypercolumn have a shared receptive field, covering the processing of all the information available within the receptive field region (Hubel and Wiesel, 1968, 1977). In our model, the hypercolumn receptive fields are formed through self-organization in a data-driven manner by means of the structural plasticity mechanism. To gain insights into this process, we visualized such receptive fields over the course of training for seven randomly chosen hypercolumns while training the BCPNN model ($H_{hid} = 400$, $M_{hid} = 100$, $fanin = 78$) on the MNIST data (Fig. 6). We can observe in the first place that the receptive fields converge to a meaningful set of spatially localized patches over the input space. We obtain these spatially contiguous patches even though there are no built-in mechanisms or constraints that would help the model exploit any spatially topological information in the input data attributes. The convergence process is rather fast and after 5000 training iterations (number of training samples) there are no significant changes in the receptive field positions (connectivity swaps due to structural plasticity, Fig. 6B). The usage score computed as a part of the structural plasticity mechanism (Eq. 9) starts at zero at the beginning of training and increases over the course of training till convergence to a positive value (Fig. 6C).

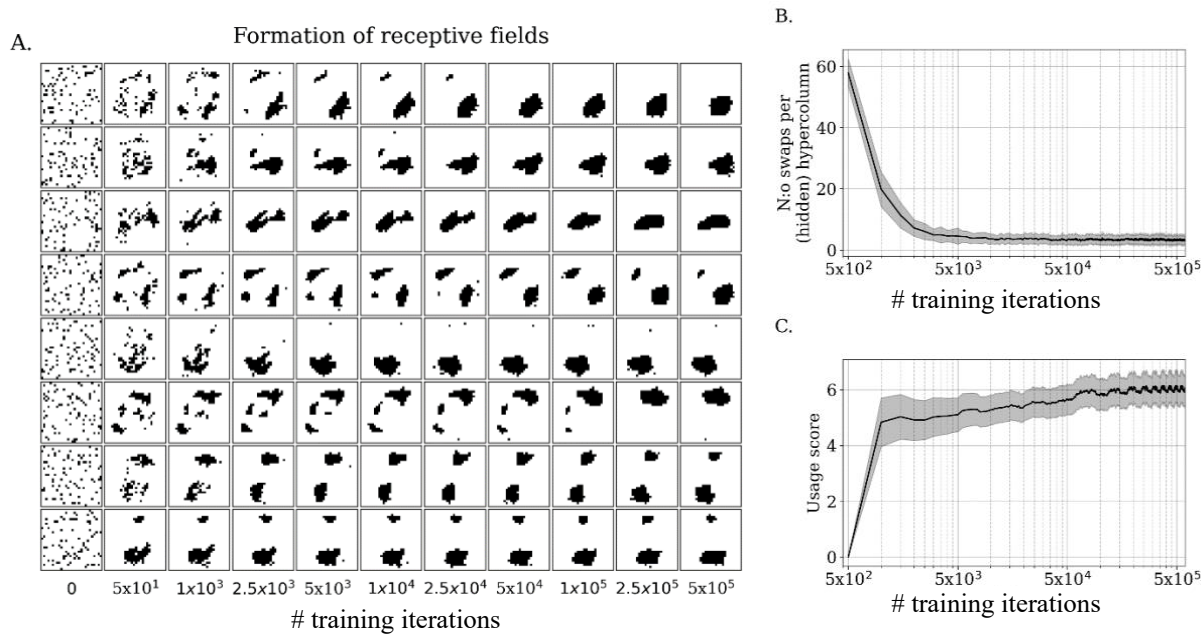

*Figure 6: Formation of hypercolumn receptive fields in the BCPNN model through structural plasticity. A. Each row corresponds to incoming connections over the course of training on MNIST dataset from the input layer to one randomly chosen hypercolumn in the hidden layer, i.e., the receptive field of that hypercolumn. The evolution of receptive fields over training iterations (samples) is shown along the horizontal direction (left to right). The connections are initially random, connecting to random pixels in the input layer. Over the training iterations, the input-to-hidden-layer connections converge to spatially localized patches of input pixels, reflecting high correlations typically found with nearby pixels in images. B. The number of swaps per hidden hypercolumn converges over the course of structural plasticity training. C. The usage score per hidden hypercolumn is maximized (starting at zero) and converges to positive value reflecting the normalized mutual information between the hidden hypercolumn and the input hypercolumns within the corresponding receptive field.*

We also evaluated the quantitative effect of structural plasticity on representation learning by measuring the linear separability of the resulting representations (linear classification accuracy). We then compared the model with the corresponding feedforward BCPNN model but with random connections, i.e., without structural plasticity (fixing the same $fanin = 78$), and with pre-defined local receptive fields (9x9 square filters; only for image datasets). The results for all 11 datasets (Table 6) clearly demonstrate the advantage that the structural plasticity offers to the BCPNN model when compared to random connectivity (p<0.001, Wilcoxon signed-rank test). Model with structural plasticity was similar in performance, even performs slightly better, compared to model with predefined locally connected model. This might be since they allocate their receptive fields more densely in the center of the image where the object is typically located, compared to the image boundaries.

*Table 6: Comparison of feedforward BCPNN model with and without structural plasticity*

| Dataset | | Random conns. | Local pre-defined conns. | Structural plasticity |
|---|---|---|---|---|
| MNIST | | 96.3 ± 0.09 | 98.5 ± 0.02 | 98.6 ± 0.06 |
| F-MNIST | | 85.6 ± 0.08 | 86.7 ± 0.05 | 88.9 ± 0.04 |
| SVHN | Grayscale DoG | 76.5 ± 0.18 | 79.2 ± 0.06 | 83.9 ± 0.14 |
| | Color DoG | 81.0 ± 0.18 | 79.5 ± 0.05 | 83.6 ± 0.18 |
|  | GMM | $76.9 \pm 0.13$ | $81.8 \pm 0.08$ | $82.2 \pm 0.21$ |
| CIFAR-10 | Grayscale DoG | $46.3 \pm 0.21$ | $51.8 \pm 0.11$ | $53.7 \pm 0.25$ |
|  | Color DoG | $53.3 \pm 0.10$ | $55.2 \pm 0.14$ | $61.2 \pm 0.15$ |
|  | GMM | $53.8 \pm 0.21$ | $56.0 \pm 0.09$ | $60.2 \pm 0.13$ |
| HIV |  | $0.80 \pm 0.08$ | N.A. | $0.83 \pm 0.09$ |
| MUV |  | $0.72 \pm 0.03$ | N.A. | $0.75 \pm 0.02$ |
| EMBER |  | $93.4 \pm 0.29$ | N.A. | $93.5 \pm 0.34$ |

## 6.3 Formation of modular sets of feature detectors through Hebbian-Bayesian learning

As shown in Section 6.2, the structural plasticity forms effective receptive fields for hypercolumns. We wanted to visualize the feature detection properties of minicolumns within their receptive fields in terms of their response characteristics. For this, we selected three random hypercolumns in the hidden layer and picked four random minicolumns from each of the chosen hypercolumns. We ran the trained BCPNN network with the test dataset and recorded all test data samples that activated the given minicolumn above a sufficiently large threshold ($\pi_{jk} > 0.9$). The results are shown in Fig. 7 for the feedforward BCPNN model when trained on MNIST (left) and F-MNIST (right) data. As can be observed from the input image samples that activated a given minicolumn, each minicolumn was activated consistently by a specific and unique feature of the image within their corresponding receptive field. Cumulatively, the hypercolumn module formed a dictionary of feature detectors, each capturing unique multivariate features within the same receptive field.

In terms of coding schemes, local representations are typically obtained with units exclusively activated by prototype matching with the data while distributed representations have units activated by unique features of the data. With this distinction in mind, the representations in feedforward BCPNN model can be more precisely characterized. The hypercolumn modules form a distributed representation with their receptive fields covering the input space. Within each hypercolumn, however, the minicolumns extract features resembling prototypes of the input within the receptive field determined by the hypercolumn and form local representations. This suggests that the feedforward BCPNN representations are "modular", where, within the hypercolumn, the minicolumns form local representations, and across hypercolumns, the minicolumns form distributed representations.

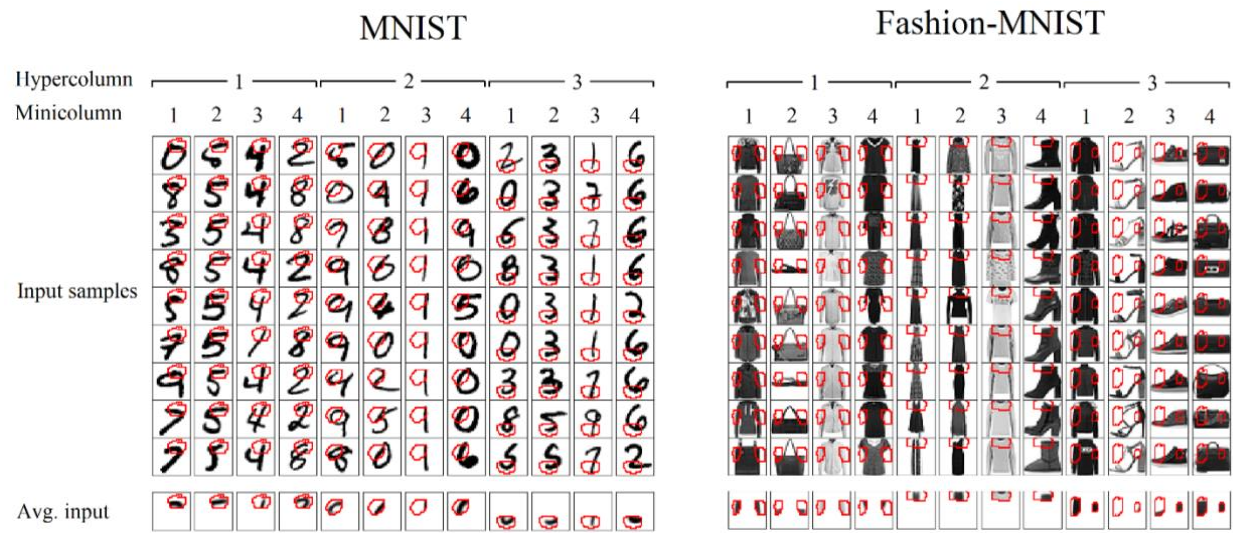

**MNIST**

**Fashion-MNIST**

*Figure 7: Features extracted by the feedforward BCPNN model trained on MNIST (left) and F-MNIST (right). Each column indicates the inputs that activated one minicolumn in the hidden layer: randomly selected samples from the test dataset (top rows) and the average of all test dataset samples (bottom row). The minicolumns are grouped by their respective hypercolumn with red boundaries indicating their respective receptive field. The minicolumns are activated consistently by inputs with the same features within their corresponding receptive field (indicated by red boundaries).*

## 6.4 Can the trained representations predict class labels?

We comparatively evaluated linear class separability of the representations learned in a purely unsupervised manner by feedforward BCPNN with those obtained with other models. To this end, we trained a linear classifier on the hidden layer representations of each model and predicted the class label on the unused test set. We report the test accuracies when trained on MNIST (Table 7), F-MNIST (Table 8), SVHN (Table 9), CIFAR-10 (Table 10), and EMBER (Table 11). For the two QSAR datasets it is more common to report the AUC-ROC scores considering the heavy imbalance of data samples across labels (Wu et al., 2018). We followed previous work and computed the AUC-ROC scores for the HIV (Table 12) and MUV (Table 13) datasets. We report the accuracy of models with all population coding methods in Appendix C. The mean accuracies and their standard deviations were estimated from $n = 5$ randomized runs.

Across all 7 datasets the average performance of feedforward BCPNN ($H_{hid} = 400$, $M_{hid} = 100$, $fanin = 78$) was consistently better than the baseline linear classifier (p=0.0156, Wilcoxon signed-rank test, $N$=7). Hence, the representations learnt by the feedforward BCPNN model in an unsupervised manner are more linearly separable (better at predicting class labels) than the input data itself. Furthermore, the feedforward BCPNN performs better on average than other unsupervised Hebbian-like learning models, AE and RBM (p=0.0156, Wilcoxon signed-rank test, $N$=7). A comparison with bio-backprop models was possible only for two datasets and it was inconclusive. For more challenging CIFAR-10 dataset, bio-backprop models gain advantage likely due to the supervised nature of the learning algorithm. Compared with DNN models, the feedforward BCPNN model is on par with 3-layer MLP in all the datasets (p=0.5625, Wilcoxon signed-rank test, $N$=7). The accuracy is relatively lower for more complex datasets like CIFAR-10 when compared to simpler datasets like MNIST, in line with previous works regarding biologically plausible models in general (Bartunov et al., 2018).

*Table 7: Comparison of linear classification performance on MNIST dataset*

| Type | Model | Accuracy [%] |
|---|---|---|
| Baseline | Linear classifier | $91.3 \pm 0.11$ |
| Unsupervised Hebbian-like learning | Feedforward BCPNN | $98.6 \pm 0.06$ |
| | RBM | $98.2 \pm 0.06$ |
| | AE | $98.3 \pm 0.03$ |
| | *l*-PCA (Illing et al., 2019) | $98.2 \pm 0.02$ |
| | *l*-ICA (Illing et al., 2019) | $98.8 \pm 0.03$ |
| | *l*-RG (Illing et al., 2019) | $98.9 \pm 0.05$ |
| | KH (Krotov and Hopfield, 2019) | $98.5 \pm 0.0$ |
| Supervised bio-backprop | BDSP (Payeur et al., 2021) | 98.9 |
| | *l*-DTP (Bartunov et al., 2018) | 98.5 |
| | *l*-FA (Bartunov et al., 2018) | 98.2 |
| DNN | 3-layer MLP | $98.6 \pm 0.03$ |
| | Homogenous Vector Capsules (Byerly et al., 2020) | 99.83 |

*Table 8: Comparison of classification performance on F-MNIST dataset*

| Type | Model | Accuracy [%] |
|---|---|---|
| Baseline | Linear classifier | $83.5 \pm 0.08$ |
| Unsupervised Hebbian-like learning | Feedforward BCPNN | $88.9 \pm 0.04$ |
| | RBM | $88.3 \pm 0.05$ |
| | AE | $87.6 \pm 0.11$ |
| DNN | 3-layer MLP | $89.5 \pm 0.14$ |

*Table 9: Comparison of classification performance on SVHN dataset*

| Type | Model | | Accuracy [%] |
|---|---|---|---|
| Baseline | Linear classifier | GMM | $30.4 \pm 0.07$ |
| Unsupervised Hebbian-like learning | Feedforward BCPNN | Grayscale DoG | $83.9 \pm 0.14$ |
| | RBM | Grayscale DoG | $72.8 \pm 0.18$ |
| | AE | GMM | $64.5 \pm 0.09$ |
| DNN | 3-layer MLP | Grayscale DoG | $82.5 \pm 0.14$ |

*Table 10: Comparison of classification performance on CIFAR-10 dataset*

| Type | Model | | Accuracy [%] |
|---|---|---|---|
| Baseline | Linear classifier | GMM | $41.9 \pm 0.08$ |
| Unsupervised Hebbian-like learning | Feedforward BCPNN | Color DoG | $61.2 \pm 0.15$ |
| | RBM | GMM | $54.2 \pm 0.23$ |
| | AE | GMM | $47.4 \pm 0.11$ |
| | *l*-PCA (Illing et al., 2019) | | $50.8 \pm 0.3$ |
| | *l*-ICA (Illing et al., 2019) | | $53.9 \pm 0.3$ |
| | *l*-RG (Illing et al., 2019) | | $52.0 \pm 0.4$ |

| | KH (Krotov and Hopfield, 2019) | | 50.8 |
|---|---|---|---|
| Supervised bio-backprop | BDSP (Payeur et al., 2021) | | 79.9 |
| | l-DTP (Bartunov et al., 2018) | | 60.5 |
| | l-FA (Bartunov et al., 2018) | | 62.6 |
| DNN | 3-layer MLP | Raw | $53.7 \pm 0.2$ |
| | Vision Transformer (Dosovitskiy et al., 2020) | | 99.5 |

*Table 11: Comparison of classification performance on EMBER dataset*

| Type | Model | | Accuracy [%] |
|---|---|---|---|
| Baseline | Linear classifier | GMM | $89.43 \pm 0.12$ |
| Unsupervised Hebbian-like learning | Feedforward BCPNN | GMM | $93.52 \pm 0.34$ |
| | RBM | GMM | $88.74 \pm 0.19$ |
| | AE | GMM | $90.13 \pm 0.22$ |
| DNN | 3-layer MLP | GMM | $94.06 \pm 0.25$ |

*Table 12: Comparison of classification performance on HIV dataset*

| Type | Model | AUC- ROC |
|---|---|---|
| Baseline | Linear classifier | $0.76 \pm 0.01$ |
| Unsupervised Hebbian-like learning | Feedforward BCPNN | $0.83 \pm 0.09$ |
| | RBM | $0.72 \pm 0.02$ |
| | AE | $0.79 \pm 0.01$ |
| DNN | 3-layer MLP | $0.77 \pm 0.01$ |
| Domain-specific models | XGBoost | 0.84 |
| | GraphConv | 0.79 |

*Table 13: Comparison of classification performance on MUV dataset*

| Type | Model | AUC-ROC |
|---|---|---|
| Baseline | Linear classifier | $0.71 \pm 0.01$ |
| Unsupervised Hebbian-like learning | Feedforward BCPNN | $0.75 \pm 0.03$ |
| | RBM | $0.69 \pm 0.01$ |
| | AE | $0.72 \pm 0.01$ |
| DNN | 3-layer MLP | $0.79 \pm 0.00$ |
| Domain-specific models | XGBoost | 0.72 |
| | GraphConv | 0.77 |

## 6.5    Hebbian-Bayesian learning converges in fewer training iterations.

It is desirable to have the learning mechanism require as few training iterations as possible for the convergence of representations. We compared the number of training iterations required for convergence of the representation learning process in the feedforward BCPNN model with other models. For this, we paused the learning at regular intervals (in log-linear scale) over the course of training and evaluated the feedforward-driven hidden representations using a linear classifier on the validation set. Each training iteration in the representation learning phase of BCPNN implies here one run through the network for one training sample (Fig. 2). Then in the evaluation step (supervised) we held out 10% of the entire training set as a validation set, trained the linear classifier on the remaining 90% of the training dataset, and recorded the validation set accuracy. We used $N = 3000$ hidden units for the other models. For a fair comparison, we used two feedforward BCPNN models: one with the same number of hidden units as in all the other models (BCPNN 30x100, $H_{hid} = 30$, $M_{hid} = 100$, $fanin = 78$) and the other with roughly the same number of trainable parameters (weights and biases) as the other models (BCPNN 400x100, $H_{hid} = 400$, $M_{hid} = 100$, $fanin = 78$).

Fig. 8 shows the trajectory of validation accuracy (mean and standard deviation over $n = 10$ randomized runs) over the course of training for all the models. Both feedforward BCPNN models clearly reached high accuracy faster than all the other models. To reach 98%, feedforward BCPNN model required around $10^4 - 10^5$ iterations (around 1-2 epochs), while other models took $10^6$ iterations (20 epochs) or more.

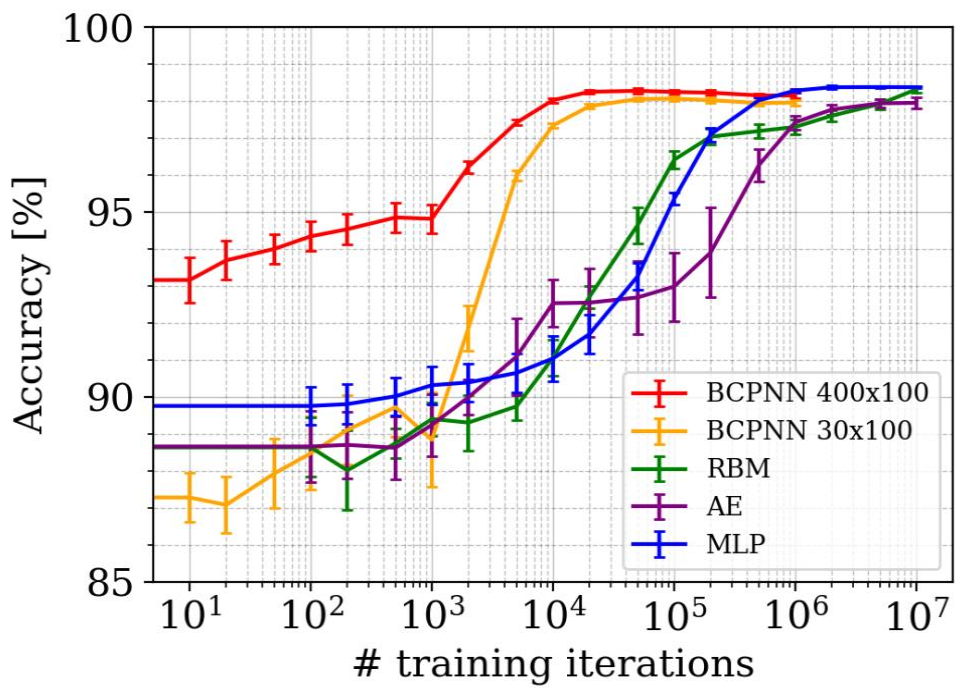

**Figure 8: Comparison of convergence of learning.** *Validation accuracy of linear classification of the hidden representations over the course of learning on MNIST dataset by the feedforward BCPNN, MLP, RBM and AE. Mean and standard deviation over 10 randomized runs.*

## 6.6    How does the network architecture influence performance?

There are three key model parameters that determine the BCPNN network architecture: (1) the total number of minicolumns in the hidden layer, $N_{hid}$ ($= H_{hid} * M_{hid}$), (2) number of minicolumns per hypercolumn in the hidden layer, $M_{hid}$, and (3) the $fanin$ of the active incoming connections from the input layer to each hidden hypercolumn. It should be noted that the $fanin$ parameter is directly linked with the connectivity percentage, as follows: $connectivity \% = 100 * fanin / H_{inp}$.

We wanted to examine the sensitivity of the model's representation capability to these architecture parameters and establish an empirical "rule of thumb" about the network configuration. To this end, we used a linear classifier and evaluated the linear separability of the representations obtained with the model configured with all possible combinations of the following parameter values: $N_{hid} \in \{1000, 10000, 100000\}$, $M_{hid} \in \{10, 20, 50, 100, 200, 500, 1000\}$ and, $fanin \in \{0.01, 0.02, 0.05, 0.1, 0.2, 0.5\} * H_{inp}$. For each parameter setting, we trained the feedforward BCPNN model on MNIST, F-MNIST, SVHN, CIFAR-10, EMBER datasets and ran $n = 5$ runs with randomized seeds.

The test results obtained with the model trained on the MNIST dataset when systematically varying each one of $N_{hid}$, $M_{hid}$, and $fanin$ parameters, while keeping the other two constant, are reported in Fig. 9A-C. More detailed results from all combination of parameter values and for all the datasets are shown in Appendix D. We observed the following trends across the majority of network parameter settings: (1) The performance was consistently better with higher $N_{hid}$ (Fig. 9A), (2) the performance when varying $M_{hid}$ followed a U-curve, performing best at an intermediate value around $M_{hid} \approx 200$ (Fig. 9B), and (3) the performance when varying $fanin$ also followed a U-curve performing best at an intermediate value around $fanin \approx 50$ (Fig. 9C).

To quantitatively verify if these parameter settings are consistently suitable across all the network configurations and across all the datasets, we calculated the distribution of optimal parameter values (Fig. 9D-F). For this, for each of the three parameters (fixing other parameters and dataset), we found the values that produced the classification error ($n = 5$ runs) within three standard deviations from the lowest classification error. We refer to this set of optimal parameters as $N_{hid}^*$, $M_{hid}^*$, and $fanin^*$ (red stars in Fig. 9A-C for MNIST data). We then plotted the distribution of the optimal values for each parameter across all datasets and across other parameter values. For $N_{hid}^*$ (Fig. 9D), there was a clear preference for higher values. This clearly demonstrated that scaling the hidden layer by increasing the total number of minicolumns improves the performance (other parameters being constant). For $M_{hid}^*$ there was clearly unimodal distribution with a preference around $50 \leq M_{hid} \leq 200$ (Fig. 9E). The value of $M_{hid}$ controls the degree of modularity in the hidden layer (for a constant $N_{hid}$) with low $M_{hid}$ indicating a hidden layer with many small modules and high $M_{hid}$ indicating a hidden layer with few large modules. This suggests that there is a preference for a modular layer architecture for learning effective hidden representations. For $fanin^*$ (Fig. 9F) there is once again a finite range of values between $50 \leq fanin \leq 200$ where the classification performance peaks. The $fanin$ controls the degree of connection sparsity with high $fanin$ indicating high connectivity ($fanin = H_{inp}$ is 100% connectivity) and low $fanin$ indicating low connectivity. The results imply that the input-to-hidden-layer connectivity is most suitable for the hidden representations at a limited degree of sparsity.

Overall, we found that the feedforward BCPNN model performance is robust over a wide range of values for all the parameters we considered. We did not focus some other parameters such as the learning rate ($\alpha$), noise amplitude ($\lambda$), number of training epochs ($N_{usup}$), number of swaps ($N_{swap}$), and threshold ($\rho$). Based on our experience, we did not consider these parameters to be of critical importance to performance.

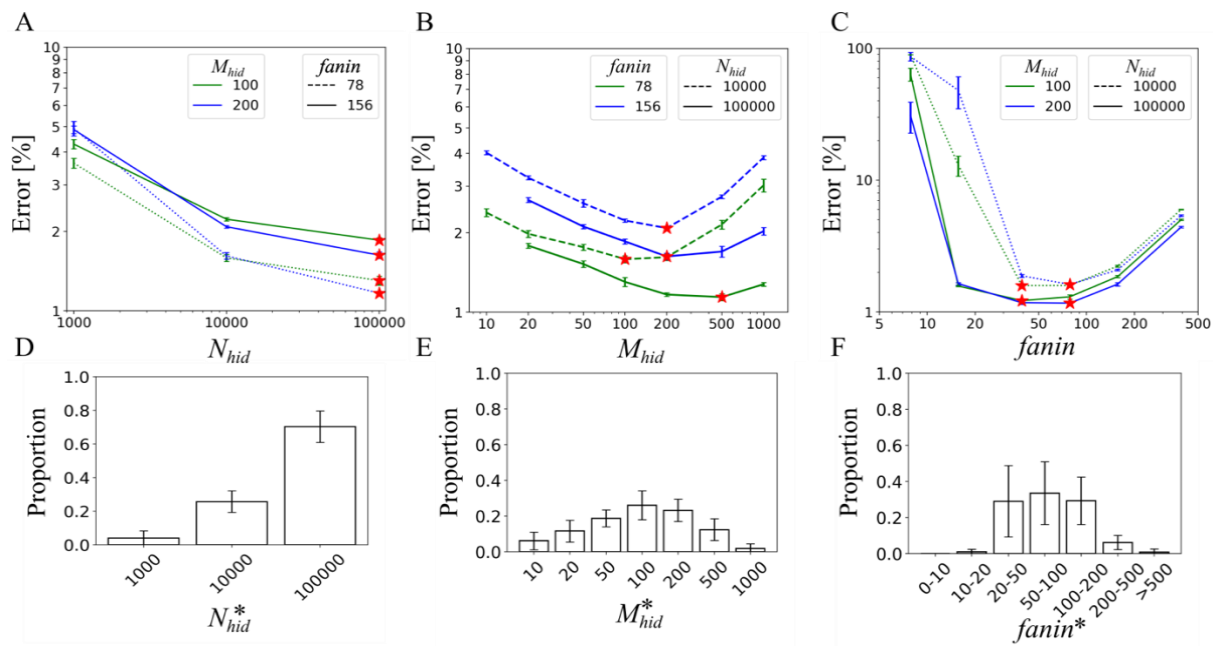

**Figure 9:** *Dependence of the network architecture of the feedforward BCPNN model on performance. We varied three parameters determining the network architecture, $N_{hid}$, $M_{hid}$, $fanin$, and trained the model on all combination of parameter values. **A-C** Results from training on MNIST dataset w.r.t to each of the three parameter values. For each of the three parameters, we found the optimal parameter values that produced the classification error within three standard deviations from the lowest classification error (indicated as red stars in figure). **D-E** Distribution of the optimal values for each parameter across all datasets and across other parameter values.*

## 7. Discussion

We have demonstrated that the feedforward BCPNN model can perform meaningful unsupervised representation learning across a wide range of datasets. Learning without any label information is considered a harder task compared to supervised learning. Additionally, learning without any global optimization algorithm as in end-to-end backprop has been considered inefficient (LeCun et al., 2015; Richards et al., 2019). Yet, our results demonstrate the feasibility and effectiveness of unsupervised representation learning using local Hebbian plasticity augmented with correlation based structural plasticity. To evaluate the linear separability, we obviously utilized class labels while training and testing the linear classifier but without affecting the unsupervised representation learning. The feedforward BCPNN model performs on par with other brain-like neural networks models and MLPs in a variety of standard machine learning benchmark problems in terms of the linear separability of learnt data representations.

It is generally regarded that learning using Hebbian plasticity, although neurobiologically motivated, is theoretically and computationally unjustified (Hinton, 1990; Richards et al., 2019). Furthermore, Hebbian plasticity employing pair-wise correlations of pre-synaptic and post-synaptic activities can be unstable as it forms a positive feedback loop with neuronal activities (larger synaptic weights drive higher activities, which in turn increase the synaptic weights) (Turrigiano, 2008a). Much effort to date has been directed at ad-hoc additional constraints that prevent Hebbian plasticity from being unstable and unbounded (Bienenstock et al., 1982a; Földiák, 1990; Oja, 1982; Turrigiano, 2008a). Connection weights under Hebbian learning are typically formulated as the linear correlation of pre- and post-synaptic activities, though this is not necessary, and might be too restrictive. The BCPNN connection weights are a monotonic non-linear function of the correlation of pre- and post-synaptic activity (Eq. 7) and is indeed strengthened when the correlation is increased, making it Hebbian learning. The BCPNN learning rule rather than being an ad-hoc modification to the correlation-based rule derives from normative principles following Bayesian inference. According to this plasticity rule, the synaptic weights increase only when the coactivations probabilities are higher than the product of individual activation probabilities. There were no additional constraints or modifications necessary for the plasticity rule to be stable, bounded and, consequently, able to learn internal representations effectively. The individual activation probabilities ($p_i$ and $p_j$ traces) in the Hebbian-Bayesian rule can be considered to implement a form of homeostatic plasticity that has been observed in the cortical synapses scaling the synaptic efficacies to maintain the neural firing rate to a constant value (Turrigiano, 2008b; Watt and Desai, 2010). Furthermore, our structural plasticity mechanism was instrumental in finding effective receptive fields for the hypercolumn modules in the hidden layer, while staying biologically plausible using local correlation-based usage scores to control swapping of activate/silent synaptic connections. We additionally found that our feedforward BCPNN model with online (incremental) learning needs far fewer training iterations compared to other brain-like neural networks and MLP, typically converging within a few epochs.

One more feature that differentiates the feedforward BCPNN model from deep learning models is the modular network architecture. Typical ANNs have layers constituting hundreds or thousands of unstructured neuron-like units receiving connections from the previous layer and sending connections to the next layer in the hierarchy. The layers in the BCPNN model, on the other hand, have layers, hypercolumns, minicolumns, and individual neurons: multiple orders of nested modular organization. The hypercolumn module, especially of interest for representation learning function, induces local competition leading to minicolumns specializing as feature detectors within their hypercolumn receptive field. Our results demonstrate the relevance of a set of brain-like attributes, such as Hebbian learning, locally competitive modular architecture, and structural plasticity, for biologically plausible backprop-free neural networks.

One of the emerging concerns for deployment of deep learning systems is the enormous energy demands for running high-performance computing clusters (Anthony et al., 2020; Henderson et al., 2020). Moreover, deep learning powered by global end-to-end backprop requires synchronization of the whole network for weight updates making them infeasible for large-scale model parallelism as well as heterogeneous and edge computing. In response to these challenges, there has been growing interest in brain-like neuromorphic algorithms that builds on more biologically plausible neural networks and holds the promise for energy-efficient neuromorphic hardware to handle real-time data streams efficiently with sparse and asynchronous event-based communication platforms (Roy et al., 2019; Schuman et al., 2022).

Previously, BCPNN-based associative memory models were shown to be efficiently implemented on the SpiNNaker neuromorphic hardware (Knight et al., 2016). BCPNN implementation on custom-designed Application-Specific Integrated Circuits (ASICs), which offers the advantage of high performance and energy efficiency with its full customized architecture, achieved three orders better efficiency compared to GPUs and SpiNNaker designed around general-purpose micro-controllers (Stathis et al., 2020). BCPNN-based associative memory models have been implemented on Field-Programmable Gate Arrays (FPGAs), which offer easy reconfigurability, high flexibility, and energy efficiency (Liu et al., 2020; Podobas et al., 2021). Memristor-based analog computing systems was shown to leverage the online localized synaptic learning of BCPNN models and offers the potential for future compute-in-memory neuromorphic hardware (Wang et al., 2021a, 2021b, 2022, 2023; Xu et al., 2021). The feedforward BCPNN model introduced in this paper was also demonstrated to be readily implemented as a spiking neural network with minimal compromise on classification performance (Ravichandran et al., 2023c, 2024) demonstrating the hardware friendly nature of the model.

A crucial step forward for the feedforward BCPNN model is to extend the architecture to multilayer (more than a single hidden layer) networks where hierarchical internal representations can be learnt from the data (Ravichandran et al., 2023b). Such multilayer "deep" architecture would allow these models to scale to large datasets such as ImageNet, which still poses a challenge for brain-like networks (Bartunov et al., 2018). The feedforward architecture can also be combined with recurrent connections projecting within each layer (Ravichandran et al., 2023). These dense recurrent connections within each cortical area have been considered the bedrock of cortical information processing (Douglas and Martin, 2007). The recurrent connections when combined with feedforward connected network in the BCPNN model (modelled as neocortical L2/3 and L4 layers, respectively) were shown to implement associative memory with attractor dynamics. While the feedforward connections extract sparse distributed high-dimensional representations, the recurrent connections perform figure-ground segmentation, robustness against occlusion, prototype extraction, and pattern completion.

From our perspective, the overarching goal of brain-like computing is two-fold. Firstly, we expect a growing demand for brain-like systems in a wide range of real-world applications with diverse implementation targets from edge computing devices to large-scale brain-like AI systems. Secondly, we consider such brain-like neural networks as simplistic but scalable models of brain function, which help us gain more insights into generic computational principles underlying the brain's perceptual and cognitive capabilities. Although our model does not incorporate nor is constrained by all available biophysical details, it still sheds some light on the functional organization of, e.g., sensory processing streams in the cortex, and how synaptic and structural plasticity could support learning in the brain. It can, most importantly, suggest what principles are likely to be functionally relevant and what biological details can be abstracted away. We envision this as the building block towards brain-like architecture for next-generation of machine intelligence.

# 8. Declaration of competing interests

The authors declare that they have no competing financial interests or personal relationships that could have appeared to influence the work reported in this paper.

# 9. Data availability

The code used to run the experiments is available publicly on GitHub and can be accessed at https://github.com/nbrav/BCPNNSim-ReprLearn.

# 10. Author contributions

A.L., N.R., and P.H. conceived the project, designed the experiments, wrote the manuscript. A.L. contributed the structural plasticity algorithm. A.L. and N.R., developed software. N.R. performed and validated the experiments, made visualizations, wrote the initial draft. P.H. and A.L. supervised the project. P.H. acquired funding for the project.

# 11. Acknowledgements

Funding for the work is received from the Swedish e-Science Research Centre (SeRC), Vetenskapsrådet (Swedish Research Council) grants no. 2018-05360, Indo-Swedish joint network 2018 grant No. 2018-07079, and European Commission Directorate-General for Communications Networks, Content and Technology grant no. 101135809 (EXTRA-BRAIN). The authors gratefully acknowledge the HPC RIVR consortium (www.hpc-rivr.si) and EuroHPC JU (eurohpc-ju.europa.eu) for funding this research by providing computing resources of the HPC system Vega at the Institute of Information Science (www.izum.si). Part of the simulations were also performed on resources provided by Swedish National Infrastructure for Computing (SNIC) at the PDC Center for High Performance Computing, KTH Royal Institute of Technology (www.pdc.kth.se). We would further like to thank Martin Rehn (KTH), Jing Gong, Artem Zhmurov, and Lilit Axner (EuroCC National Competence Center Sweden) for collaborating and developing OpenACC and CUDA implementation of the model on GPUs.

# 12. Appendix A. Feedforward BCPNN maximizes data likelihood

The computations involved in the BCPNN model converts probabilistic inference rules into neural computation. For the feedforward BCPNN model, each hypercolumn in the hidden layer optimizes the input layer's data likelihood following the Expectation-Maximization (EM) procedure (within its receptive field formed by structural plasticity). The neuronal activation and synaptic plasticity rules can be normatively formulated as iterative updates of the EM algorithm and correspond to the E and M steps respectively.

We consider a dataset constituting $N_{sample}$ training samples, with each training iteration of the model (indexed by $n$) processing one data sample independently drawn from the dataset. The data samples constitute a set of $H_{inp}$ attributes, $\bar{X} = \{X_1, \dots, X_{H_{inp}}\}$, with each attribute $X_i, i \in \{1, \dots, H_{inp}\}$, being a discrete one-hot coded vector, taking one of $M_{inp}$ feature values, $X_i = x_{im}, m \in \{1, \dots, M_{inp}\}$. This condition is satisfied if the dataset is already discrete-coded, otherwise, we first used the input layer to convert continuous-coded data to discrete-coded population vector representations and use this as the input. Also, we here assumed that every hypercolumn in a layer to have the same $M_{inp}$, but this is not necessary theoretically and is used for the sake of simplicity of notational convenience.

Straightforward closed-form optimization of the likelihood is considered non-trivial due to the complex dependencies existing between the data attributes. However, when we consider the data distribution along with a hidden (latent) variable, $Y$, the joint probability is considered simpler to estimate since the hidden variable can act as the underlying factor of variation (Bengio et al., 2013). The problem of fitting the feedforward BCPNN model to the dataset can be framed as finding the parameters for the model such that the model likelihood $P(\bar{X})$ for the dataset is maximized.

The iterative EM solution is to maximize the term $F(\bar{X})$ (Dempster et al., 1977; Neal and Hinton, 1998):

$$F(\bar{X}) = \sum_Y Q(Y) \log \frac{P(Y, \bar{X})}{Q(Y)} \qquad (A1)$$

We can see that maximizing $F(\bar{X})$ maximizes the lower bound on the log likelihood (Neal and Hinton, 1998) as they follow the below relation:

$$\log P(\bar{X}) = F(\bar{X}) + D_{KL}(Q, P) \geq F(\bar{X}) \qquad (A2)$$

where $D_{KL}$ is the Kullback-Liebler divergence between distributions. The EM algorithm maximize $F(\bar{X})$ by iteratively updating the newly introduced distribution $Q$ over the hidden variable, and the joint probability distribution $P$ over the input and hidden variables. The solution consists of gradient ascent steps iteratively over both $Q$ and $P$ which yields the following updates:

$$\log Q^{(n)}(Y) = \log P^{(n)}(Y, \bar{X}) + c \qquad \text{(E step; A3)}$$

$$P^{(n+1)}(Y, \bar{X}) \propto Q^{(n)}(Y) \qquad \text{(M step; A4)}$$

The joint probability of a data vector can be factorized into the prior probability and a product of likelihoods of each data attribute as follows:

$$P(Y, \bar{X}) = P(Y) \prod_{i=1}^{H_{inp}} P(X_i | Y) \qquad (A5)$$

Since all the BCPNN attributes are discrete variables, we can define the following probabilities:

$$P(Y) = \prod_{k=1}^{M_{hid}} p_k^{y_k}, \qquad P(X_i | Y = y_k) = \prod_{m=1}^{M_{inp}} p_{imk}^{x_{im}} \qquad (A6)$$

where $p_k = P(Y = y_k)$ and $p_{imk} = P(X = x_{im} | Y = y_k)$. Substituting the terms of Eq. A6 in Eq. A5, the log of the joint probability can be written as:

$$\log P(Y, \bar{X}) = \sum_{k=1}^{M_{hid}} y_k \left( \log p_k + \sum_{i=1}^{H_{inp}} \sum_{m=1}^{M_{inp}} x_{im} \log p_{imk} \right) \qquad (A7)$$

Now, the E-step updates (Eq. A3) can be written for the above joint probability (Eq. A7) as follows:

$$s_k^{(n)} \stackrel{\text{def}}{=} b_k^{(n)} + \sum_{i=1}^{H_{inp}} \sum_{m=1}^{M_{inp}} \pi_{im}^{(n)} w_{imk}^{(n)}, \qquad (A8)$$

$$\pi_k^{(n)} \overset{\text{def}}{=} \frac{\exp s_k^{(n)}}{\sum_{l=1}^{M_{hid}} \exp s_l^{(n)}}. \tag{A9}$$

where $s_k^{(n)}$ is the unnormalized log probability acting as the support term, from which the probability $Q^{(n)}(Y = y_k)$ can be recovered as the minicolumn activities $\pi_k^{(n)}$ through softmax normalization. The M-step updates (Eq. A4) can be written by maximizing the following terms:

$$b_k^{(n+1)} \overset{\text{def}}{=} \log p_k^{(n)}$$

$$w_{imk}^{(n+1)} \overset{\text{def}}{=} \log \frac{p_{imk}^{(n)}}{p_{im}^{(n)} p_k^{(n)}}$$

Hence, the E step and M step of maximum likelihood procedure corresponds to the neuronal activation rule and synaptic learning rule of feedforward BCPNN model with one hidden hypercolumn.

There are other brain-like models that perform learning by maximizing the data likelihood using the EM algorithm. The Restricted Boltzmann Machine (RBM) is one such neural network model where a layer of visible units (similar to our data layer) is connected to a layer of hidden units (our hidden layer). Learning in the RBM maximizes the data likelihood using the Contrastive Divergence (CD) learning algorithm. CD learning iteratively propagates activities forward and backward between the visible and hidden layers using symmetric bidirectional weights which is not biologically plausible. In contrast, BCPNN uses unidirectional connection weights and requires running the network only once in the feedforward direction. The BCPNN model also uses hypercolumn modules with softmax activations while RBM uses binary units with stochastic sigmoid activations.

# 13.   Appendix B. Pseudocode for Representation Learning

**ALGORITHM.** UNSUPERVISED LEARNING
**input:** training dataset, dimension $N_{train}$ x $H_{inp}$ x $M_{inp}$
**for** $e$ = 1 to $N_{usup}$: # epoch
    **for** $p$ = 1 to $N_{train}$: # training sample
        **set** sample, dimension $H_{inp}$ x $M_{inp}$, as input activities
        UPDATE ACTIVITIES
        SYNAPTIC PLASTICITY
        **if** $p$ % $T_{swap}$ = 0 ?
            STRUCTURAL PLASTICITY
        **end if**
    **end for**
**end for**

**FUNCTION** UPDATE ACTIVITIES
**input** $\{\pi_{im}^{(n)}; \ i = \{1:H_{inp}\}, m = \{1:M_{inp}\}\}$
**for** $j$ = 1 to $H_{hid}$

    **for** $k = 1$ to $M_{hid}$

        update $s_{jk}^{(n)}$ (Eq. 1)

        update $\pi_{jk}^{(n)}$ (Eq. 2)

    **end for**

**end for**

**FUNCTION** SYNAPTIC PLASTICITY

**input** $\{\pi_{im}^{(n)};\ i = \{1{:}H_{inp}\}, m = \{1{:}M_{inp}\}\},\ \{\pi_{jk}^{(n)};\ j = 1{:}H_{hid}, k = 1{:}M_{hid}\}$

**for** $j = 1$ to $H_{hid}$

    **for** $k = 1$ to $M_{hid}$

        **for** $i = 1$ to $H_{inp}$

            **for** $m = 1$ to $M_{inp}$

                update $p_{im}^{(n)}, p_{jk}^{(n)}$, and $p_{imjk}^{(n)}$ (Eq. 3, 4, and 5 resp.)

            **end for**

        **end for**

    **end for**

**end for**

**for** $j = 1$ to $H_{hid}$

    **for** $k = 1$ to $M_{hid}$

        update $b_{jk}^{(n)}$ (Eq. 6)

        **for** $i = 1$ to $H_{inp}$

            **for** $m = 1$ to $M_{inp}$

                update $w_{imjk}^{(n)}$ (Eq. 7)

            **end for**

        **end for**

    **end for**

**end for**

**FUNCTION** STRUCTURAL PLASTICITY

**input** $\{w_{imjk}^{(n)};\ i = \{1{:}H_{inp}\}, m = \{1{:}M_{inp}\}, j = 1{:}H_{hid}, k = 1{:}M_{hid}\}$

**for** $j = 1$ to $H_{hid}$

    **for** $swapid = 1$ to $N_{swap}$

        **for** $i = 1$ to $H_{inp}$

            update $U_{ij}^{(n)}$

        **end for**

        $s \leftarrow \underset{i \in silent}{\operatorname{argmax}}\ U_{ij}^{(n)}$

        $a \leftarrow \underset{i \in active}{\operatorname{argmin}}\ U_{ij}^{(n)}$

        **if** $U_{sj}^{(n)} > \rho\ U_{aj}^{(n)}$

            $c_{sj}^{(n)} \leftarrow silent$

            $c_{aj}^{(n)} \leftarrow active$

        **end if**

    **end for**

end for

# 14. Appendix C. Full classification accuracy results

We evaluated the linear classification accuracy of the representations learned by representation learning models. We report the test accuracies when trained on MNIST (Table A1), F-MNIST (Table A2), SVHN (Table A3), CIFAR-10 (Table A4), EMBER (Table A5), HIV (Table A6), and MUV (Table A7).

Table A1: Comparison of classification performance on MNIST dataset

| Type | Model | Accuracy [%] |
|---|---|---|
| Baseline | Linear classifier | $91.3 \pm 0.11$ |
| Unsupervised Hebbian-like learning | Feedforward BCPNN | $98.6 \pm 0.06$ |
| | RBM | $98.2 \pm 0.06$ |
| | AE | $98.3 \pm 0.03$ |
| | *l*-PCA (Illing et al., 2019) | $98.2 \pm 0.02$ |
| | *l*-ICA (Illing et al., 2019) | $98.8 \pm 0.03$ |
| | *l*-RG (Illing et al., 2019) | $98.4 \pm 0.05$ |
| | KH (Krotov and Hopfield, 2019) | $98.5 \pm 0.0$ |
| Supervised bio-backprop | BDSP (Payeur et al., 2021) | 98.9 |
| | *l*-DTP (Bartunov et al., 2018) | 98.5 |
| | *l*-FA (Bartunov et al., 2018) | 98.2 |
| DNN | 3-layer MLP | $98.6 \pm 0.03$ |

Table A2: Comparison of classification performance on F-MNIST dataset

| Type | Model | Accuracy [%] |
|---|---|---|
| Baseline | Linear classifier | $83.5 \pm 0.08$ |
| Unsupervised Hebbian-like learning | Feedforward BCPNN | $88.9 \pm 0.04$ |
| | RBM | $88.3 \pm 0.05$ |
| | AE | $87.6 \pm 0.11$ |
| DNN | 3-layer MLP | $89.5 \pm 0.14$ |

Table A3: Comparison of classification performance on SVHN dataset

| Type | Model | | Accuracy [%] |
|---|---|---|---|
| Baseline | Linear classifier | Raw | $25.5 \pm 0.08$ |
| | | Grayscale DoG | $22.2 \pm 0.14$ |
| | | Color DoG | $24.8 \pm 0.17$ |
| | | GMM | $30.4 \pm 0.07$ |
| Unsupervised Hebbian-like learning | Feedforward BCPNN | Grayscale DoG | $83.9 \pm 0.14$ |
| | | Color DoG | $83.6 \pm 0.18$ |
| | | GMM | $82.2 \pm 0.21$ |
| | RBM | Raw | $63.6 \pm 0.24$ |
| | | Grayscale DoG | $72.8 \pm 0.18$ |
| | | Color DoG | $69.5 \pm 0.17$ |
| | | GMM | $64.1 \pm 0.05$ |

| | AE | Raw | $39.0 \pm 0.14$ |
|---|---|---|---|
| | | Grayscale DoG | $60.9 \pm 0.17$ |
| | | Color DoG | $56.3 \pm 0.18$ |
| | | GMM | $64.5 \pm 0.09$ |
| DNN | 3-layer MLP | Raw | $81.4 \pm 0.28$ |
| | | Grayscale DoG | $82.5 \pm 0.14$ |
| | | Color DoG | $79.5 \pm 0.23$ |
| | | GMM | $65.3 \pm 0.06$ |

Table A4: Comparison of classification performance on CIFAR-10 dataset

| Type | Model | | Accuracy [%] |
|---|---|---|---|
| Baseline | Linear classifier | Raw | $40.5 \pm 0.18$ |
| | | Grayscale DoG | $23.2 \pm 0.16$ |
| | | Opp. color DoG | $32.1 \pm 0.17$ |
| | | GMM | $41.9 \pm 0.08$ |
| Unsupervised Hebbian-like learning | Feedforward BCPNN | Grayscale DoG | $53.7 \pm 0.25$ |
| | | Color DoG | $61.2 \pm 0.15$ |
| | | GMM | $60.2 \pm 0.13$ |
| | RBM | Raw | $49.9 \pm 0.18$ |
| | | Grayscale DoG | $30.3 \pm 0.26$ |
| | | Opp. Color DoG | $38.8 \pm 0.21$ |
| | | GMM | $54.7, 53.34$ |
| | AE | Raw | $42.3 \pm 0.29$ |
| | | Grayscale DoG | $43.6 \pm 0.25$ |
| | | Color DoG | $47.4 \pm 0.11$ |
| | | GMM | $47.4 \pm 0.11$ |
| | *l*-PCA (Illing et al., 2019) | | $50.8 \pm 0.3$ |
| | *l*-ICA (Illing et al., 2019) | | $53.9 \pm 0.3$ |
| | *l*-RG (Illing et al., 2019) | | $52.0 \pm 0.4$ |
| | (Krotov and Hopfield, 2019) | | 50.8 |
| Supervised bio-backprop | BDSP (Payeur et al., 2021) | | 79.9 |
| | *l*-DTP (Bartunov et al., 2018) | | 60.5 |
| | *l*-FA (Bartunov et al., 2018) | | 62.6 |
| DNN | 3-layer MLP | Raw | $53.7 \pm 0.2$ |
| | | Grayscale DoG | 37.41 |
| | | Color DoG | 43.21 |
| | | GMM | 51.04 |

Table A5: Comparison of classification performance on EMBER dataset

| Type | Model | | Accuracy [%] |
|---|---|---|---|
| Baseline | Linear classifier | GMM | $89.43 \pm 0.12$ |
| Unsupervised Hebbian-like learning | Feedforward BCPNN | GMM | $93.52 \pm 0.34$ |
| | RBM | GMM | $88.74 \pm 0.19$ |
| | AE | GMM | |
| DNN | 3-layer MLP | GMM | $94.06 \pm 0.25$ |

Table A6: Comparison of classification performance on HIV dataset

| Type | Model | AUC- ROC |
|------|-------|----------|
| Baseline | Linear classifier | $0.76 \pm 0.01$ |
| Unsupervised Hebbian-like learning | Feedforward BCPNN | $0.83 \pm 0.09$ |
| | RBM | $0.73 \pm 0.02$ |
| | AE | $0.80 \pm 0.01$ |
| DNN | 3-layer MLP | $0.77 \pm 0.01$ |
| Domain-specific models | XGBoost | 0.84 |
| | GraphConv | 0.79 |

Table A7: Comparison of classification performance on MUV dataset

| Type | Model | AUC-ROC |
|------|-------|---------|
| Baseline | Linear classifier | $0.71 \pm 0.01$ |
| Unsupervised Hebbian-like learning | Feedforward BCPNN | $0.75 \pm 0.03$ |
| | RBM | $0.69 \pm 0.01$ |
| | AE | $0.72 \pm 0.01$ |
| DNN | 3-layer MLP | $0.79 \pm 0.00$ |
| Domain-specific models | XGBoost | 0.72 |
| | GraphConv | 0.77 |

## 15.    Appendix D. Impact of model parameters on performance

We experimented the influence of the model parameters by varying three key parameters of the feedforward BCPNN model: (1) the total number of hidden layer minicolumns, $N_{hid}$ ($= H_{hid} * M_{hid}$), (2) number of minicolumns per hidden layer hypercolumn, $M_{hid}$, and (3) the $fanin$ of the active incoming connections from the input layer to each hidden layer hypercolumn. For each parameter setting, we trained the feedforward BCPNN model on MNIST, F-MNIST, SVHN (Grayscale DoG), CIFAR-10 (Grayscale DoG and GMM), EMBER datasets and ran $n = 5$ runs with randomized seeds and calculated the classification error (%). The classification error with respect to $fanin$ followed an inverted U-shaped relationship (Fig. A1) across all values of $N_{hid}$ and $M_{hid}$ and across all datasets considered. Similarly, the classification error with respect to $M_{hid}$ followed an inverted U-shaped relationship (Fig. A2) across all values of $N_{hid}$ and $fanin$ and across all datasets considered. Lastly, the classification error with respect to $N_{hid}$ was consistently lower with lower $N_{hid}$ (Fig. A1, A2) across all values of $M_{hid}$ and $fanin$. We found the optimal parameters, $M_{hid}^*$ and $fanin^*$, that produced the classification error within three standard deviations from the lowest classification error for each setting of the other parameters (indicated with red stars in Fig. A1, A2) and used these values for calculation of the distribution of optimal parameter values (described in Section 6.6).

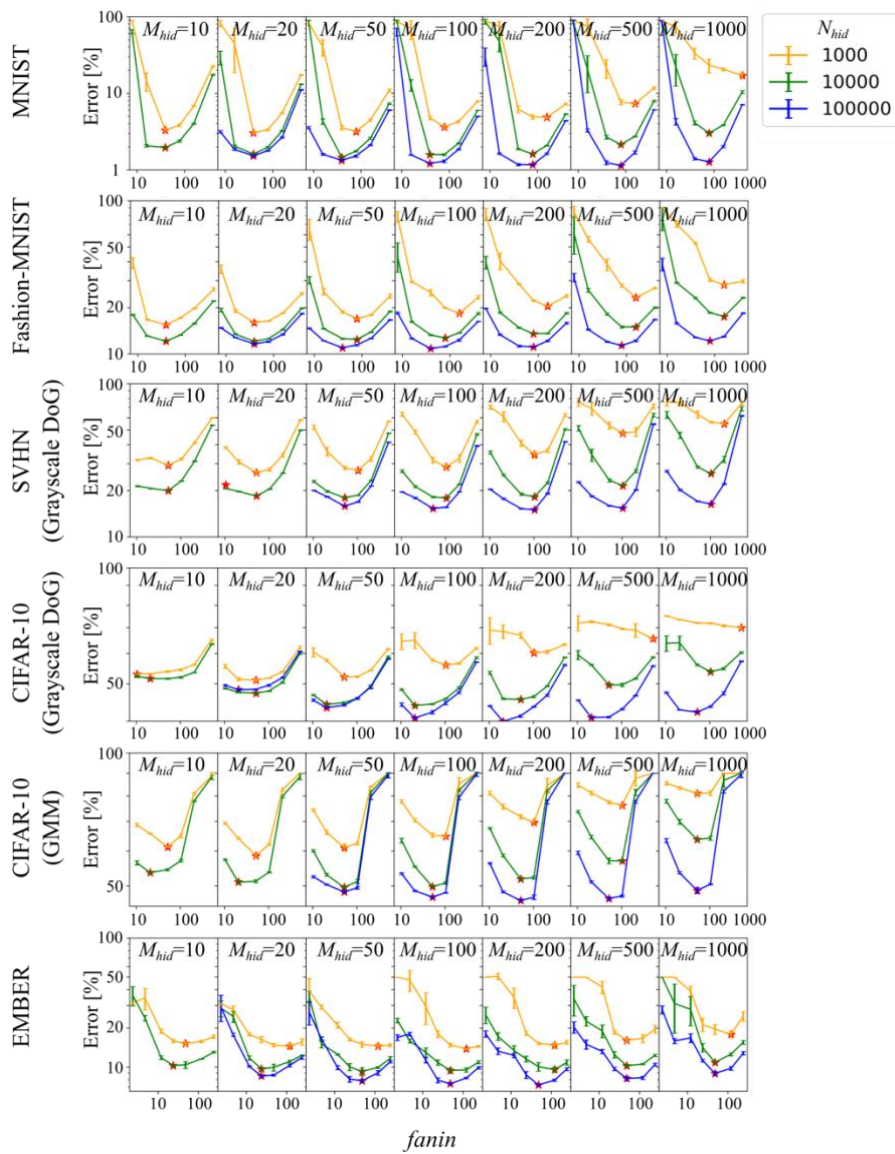

**Figure A1: Impact of** fanin **on the feedforward BCPNN model performance.** *Classification error (%) w.r.t* fanin *parameter (number of active incoming connections) shows an inverted U-shaped relationship across all values of $N_{hid}$ (ordered by color) and $M_{hid}$ (ordered column-wise) and across all datasets (ordered row-wise) considered.*

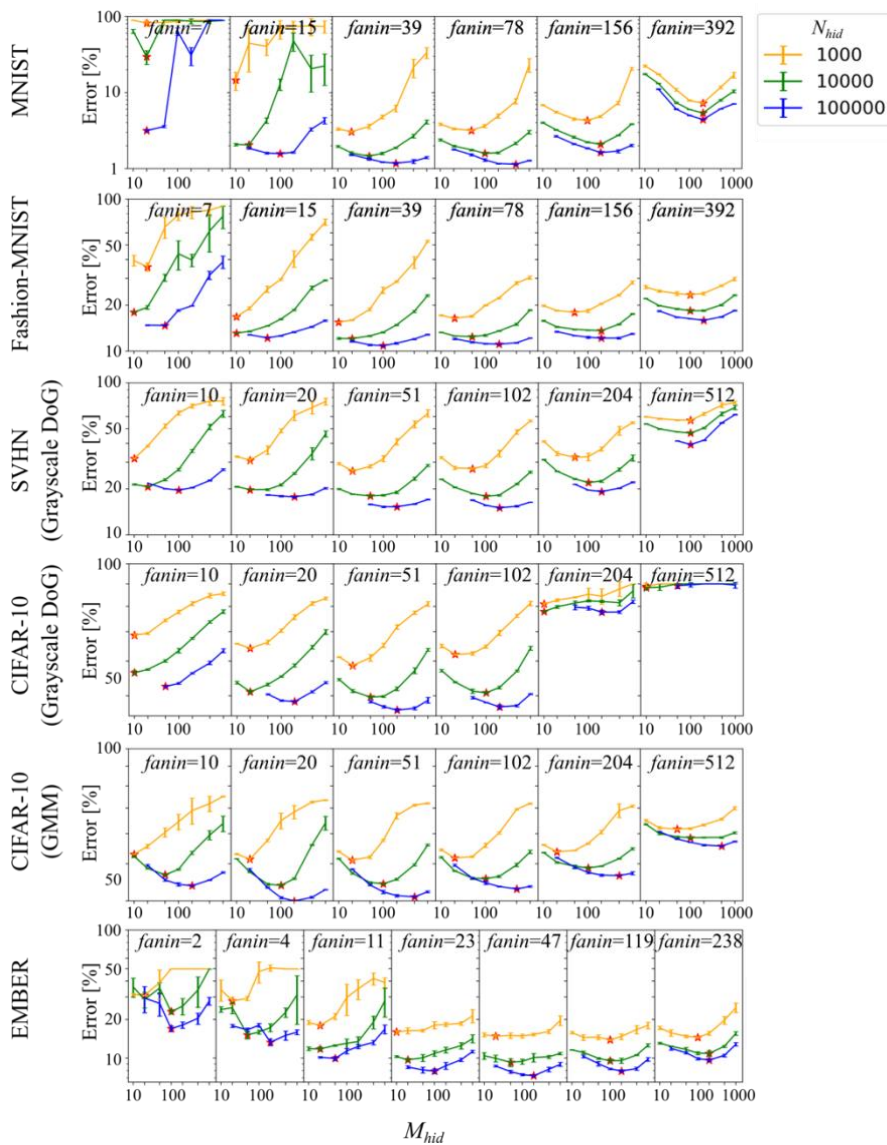

***Figure A2**: **Impact of $M_{hid}$ on the feedforward BCPNN model performance.** Classification error (%) w.r.t $M_{hid}$ parameter (number of minicolumns per hidden layer hypercolumn) shows an inverted U-shaped relationship across all values of $N_{hid}$ (ordered by color) and $fanin$ (ordered column-wise) and across all datasets (ordered row-wise) considered.*